\documentclass[nohyperref]{article}

\usepackage{microtype}
\usepackage{graphicx}
\usepackage{subfigure}
\usepackage{booktabs} 

\usepackage{hyperref}

\usepackage[accepted]{icml2023}

\usepackage{amsmath}
\usepackage{amssymb}
\usepackage{mathtools}
\usepackage{amsthm}

\usepackage[capitalize,noabbrev]{cleveref}

\theoremstyle{plain}
\newtheorem{theorem}{Theorem}[section]

\theoremstyle{definition}
\newtheorem{definition}[theorem]{Definition}

\theoremstyle{remark}

\usepackage[textsize=tiny]{todonotes}

\usepackage{xcolor}

\newcommand\CHANGE[1]{\textcolor{blue}{#1}}

\usepackage{amsmath,amsfonts,bm}

\def\eqref#1{equation~\ref{#1}}

\def\1{\bm{1}}

\DeclareMathAlphabet{\mathsfit}{\encodingdefault}{\sfdefault}{m}{sl}
\SetMathAlphabet{\mathsfit}{bold}{\encodingdefault}{\sfdefault}{bx}{n}

\def\gA{{\mathcal{A}}}

\def\gC{{\mathcal{C}}}
\def\gD{{\mathcal{D}}}

\def\gM{{\mathcal{M}}}

\def\gO{{\mathcal{O}}}
\def\gP{{\mathcal{P}}}

\def\gS{{\mathcal{S}}}
\def\gT{{\mathcal{T}}}

\def\sR{{\mathbb{R}}}

\newcommand{\E}{\mathbb{E}}

\DeclareMathOperator*{\argmax}{arg\,max}

\newcommand{\shortn}{\textup{\texttt{-}}}
\newcommand{\shorte}{\textup{\texttt{=}}}
\newcommand{\shortp}{\textup{\texttt{+}}}
\newcommand{\shortl}{\textup{\texttt{<}}}

\newcommand{\ie}{\textit{i}.\textit{e}.}
\newcommand{\eg}{\textit{e}.\textit{g}.}

\newcommand{\etc}{\textit{etc}.}

\newcommand{\name}{SARD}
\newcommand{\nameKfive}{SARD (\textit{K}=5)}
\newcommand{\nameKthree}{SARD (\textit{K}=3)}
\newcommand{\nameKone}{SARD (\textit{K}=1)}
\newcommand{\nameUnstruct}{SARD (\textit{Unstructured})}
\newcommand{\tact}{Transform2Act}
\newcommand{\nge}{NGE}

\newcommand{\originalant}{Handcrafted Robot}
\newcommand{\locomotionft}{Locomotion on Flat Terrain}
\newcommand{\locomotionvt}{Locomotion on Variable Terrain}
\newcommand{\escape}{Escape Bowl}
\newcommand{\pointnav}{Point Navigation}
\newcommand{\manipulationbox}{Manipulate Box}
\newcommand{\patrol}{Patrol}

\icmltitlerunning{Symmetry-Aware Robot Design}

\begin{document}

\twocolumn[
\icmltitle{Symmetry-Aware Robot Design with Structured Subgroups}




\begin{icmlauthorlist}
\icmlauthor{Heng Dong}{thu}
\icmlauthor{Junyu Zhang}{hust}
\icmlauthor{Tonghan Wang}{hu}
\icmlauthor{Chongjie Zhang}{wustl}


\end{icmlauthorlist}

\icmlaffiliation{thu}{Tsinghua University}
\icmlaffiliation{hust}{Huazhong University of Science and Technology}
\icmlaffiliation{hu}{Harvard University}
\icmlaffiliation{wustl}{Washington University in St. Louis}

\icmlcorrespondingauthor{Heng Dong}{drdhxi@gmail.com}
\icmlcorrespondingauthor{Chongjie Zhang}{chongjie@wustl.edu}

\icmlkeywords{Machine Learning, ICML}

\vskip 0.3in
]



\printAffiliationsAndNotice{}  

\begin{abstract}
Robot design aims at learning to create robots that can be easily controlled and perform tasks efficiently. Previous works on robot design have proven its ability to generate robots for various tasks. However, these works searched the robots directly from the vast design space and ignored common structures, resulting in abnormal robots and poor performance. To tackle this problem, we propose a Symmetry-Aware Robot Design (\name{}) framework that exploits the structure of the design space by incorporating symmetry searching into the robot design process. Specifically, we represent symmetries with the subgroups of dihedral group and search for the optimal symmetry in structured subgroups. Then robots are designed under the searched symmetry. In this way, \name{} can design efficient symmetric robots while covering the original design space, which is theoretically analyzed. We further empirically evaluate \name{} on various tasks, and the results show its superior efficiency and generalizability.
\end{abstract}

\section{Introduction}


Humans have been dreaming of creating creatures with morphological intelligence for decades~\cite{sims1994evolving, sims1994evolving3D, yuan2021transform2act, gupta2021embodied}. A promising solution for this challenging problem is to generate robots with various functionalities in simulated environments~\cite{wang2019neural, yuan2021transform2act}, in which robots' functionalities are largely determined by their designs and control policies. Learning control policies for handcrafted robots with fixed designs has been extensively studied in previous works ~\cite{schulman2017PPO, fujimoto2018TD3, huang2020one, dong2022solar}. However, as the other critical component, the design of robots has attracted scant attention and achieved limited success in the literature. The field of automatic robot design aims at searching for optimal robot morphologies that can be easily controlled and perform various tasks efficiently. This problem has been a long-lasting challenge, mainly for two reasons: 1) the design space, including skeletal structures and attributes of joints and limbs, is large and combinatorial, and 2) the evaluation of each design requires training and testing an optimal control policy which is often computationally expensive. 

For automatic robot design, prior works~\cite{gupta2021embodied, wang2019neural} typically adopt evolutionary search (ES) algorithms, where robots are sampled from a large population and learn to perform tasks independently during an iteration. At the end of each iteration, robots with the worst performance are eliminated, and the surviving robots will produce child robots by random mutations to maintain the population. Recently, \citet{yuan2021transform2act} discussed the low sample efficiency problem in ES-based methods, \eg{}, robots in the population do not share their training experiences and zeroth-order optimization methods such as ES are sample-inefficient for high-dimensional search space~\cite{vemula2019contrasting}. They used reinforcement learning (RL) to sample and optimize robot designs by incorporating the design process into the control process and sharing the design and control policies across all robots.

\begin{figure}[t]
\centering
\includegraphics[width=\linewidth]{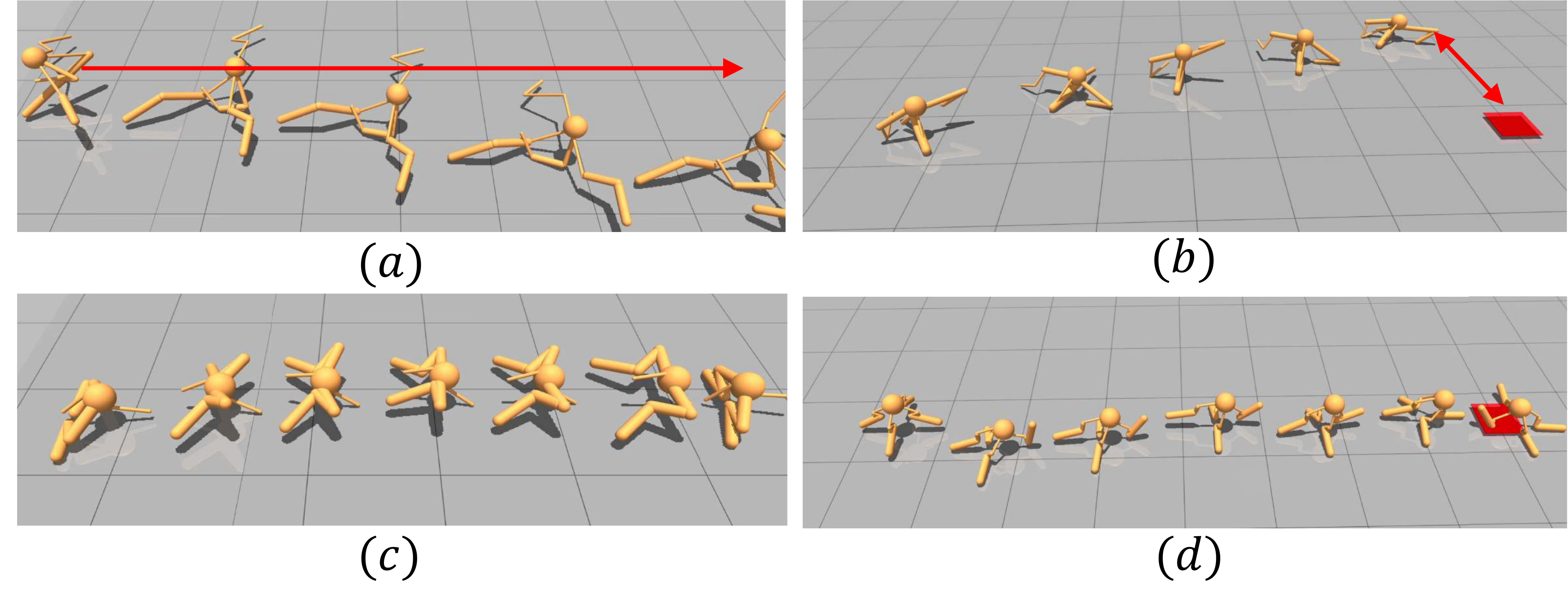}
\vspace{-2em}
\caption{Time-lapse images of robots with different symmetries performing various tasks: (a, c) running forward; (b, d) reaching random goals. Robots designed by prior work~\cite{yuan2021transform2act} do not satisfy any non-trivial symmetry and might be hard to control: (a) the robot deviated from the right direction; (b) the robot missed the goal. Different tasks may require different symmetries: (c) bilateral symmetry is suitable for tasks involving only running forward; (d) radial symmetry for reaching random goals.} \label{fig:control}
\vspace{-1em}
\end{figure}

Despite the progress, robots designed by these approaches are intuitively abnormal, empirically hard to control, and ultimately result in poor performance. Examples are provided in the time-lapse images in \cref{fig:control} (a-b), in which robots designed by prior work~\cite{yuan2021transform2act} perform poorly on different tasks. We hypothesize that the underperformance can be attributable to the fact that most prior works directly search for robots in the whole vast design space without exploiting useful structures that can largely reduce the search space. 

To verify this hypothesis, in this paper, we explore utilizing \emph{symmetry} as the key characteristic to unveil the structure of the design space and hereby reduce learning complexity. Symmetry is one structure commonly observed in biological organisms~\cite{savriama2011symmetry}, \eg{}, bilateral symmetry in flies~\cite{evans2010axon}, radial symmetry in jellyfish~\cite{abrams2015self}, and spherical symmetry in bacteria~\cite{shao2017growth}. From the learning perspective, symmetry-aware robot design has two advantages. First, it requires searching for much fewer robot designs. If one design turns out to be unsuitable for the current task, other designs from the same symmetry can be searched less frequently as they are likely to be morphologically and functionally similar. Second, symmetric designs can reduce the degree of control required to learn balancing~\cite{raibert1986symmetry, raibert1986legged} as in \cref{fig:control} (a)(c). Prior works noticed the benefits of symmetry~\cite{gupta2021embodied,wang2019neural}, but only bilateral symmetry was considered. Other tasks may require different symmetries. For example, tasks that involve running in different directions require radial symmetry (\cref{fig:control} (d)). However, none of the previous works explored learning suitable robot symmetries for different tasks.

In this paper, we introduce a novel Symmetry-Aware Robot Design (\name{}) framework. Realizing this framework involves two major challenges. The first challenge is how to represent symmetries and how to find optimal symmetry. To consider a wide range of symmetries while also avoiding extra learning complexity, we propose to use the subgroups of dihedral group~\cite{gallian2021contemporary} to represent symmetries. Each subgroup represents a kind of symmetry and a symmetric space. The trivial subgroup containing only the identity group element is exactly the original design space. To find the optimal symmetry efficiently, we utilize the group structures and adopt a simple local search algorithm applied in the structured subgroups for smoothly changing symmetry types to alleviate gradient conflict problem~\cite{liu2021conflict}. The second challenge is how to design robots that satisfy a given symmetry. We propose a novel plug-and-play symmetry transformation module to map any robot design into a given symmetric space. We also provide theoretical analysis to verify that the transformed robot designs are in the given symmetric space and this module can cover the whole space.

We evaluate our \name{} framework on six MuJoCo~\cite{todorov2012mujoco} tasks adapted from~\citet{gupta2021embodied}. \name{} significantly outperforms previous state-of-the-art algorithms in terms of both sample efficiency and final performance. Performance comparison and the visualization of the symmetry learning process strongly support the effectiveness of our symmetry searching and transformation approaches. Our experimental results highlight the importance of considering various symmetries in robot design.

\section{Related Works}

\begin{figure*}[htp]
\centering
\includegraphics[width=\linewidth]{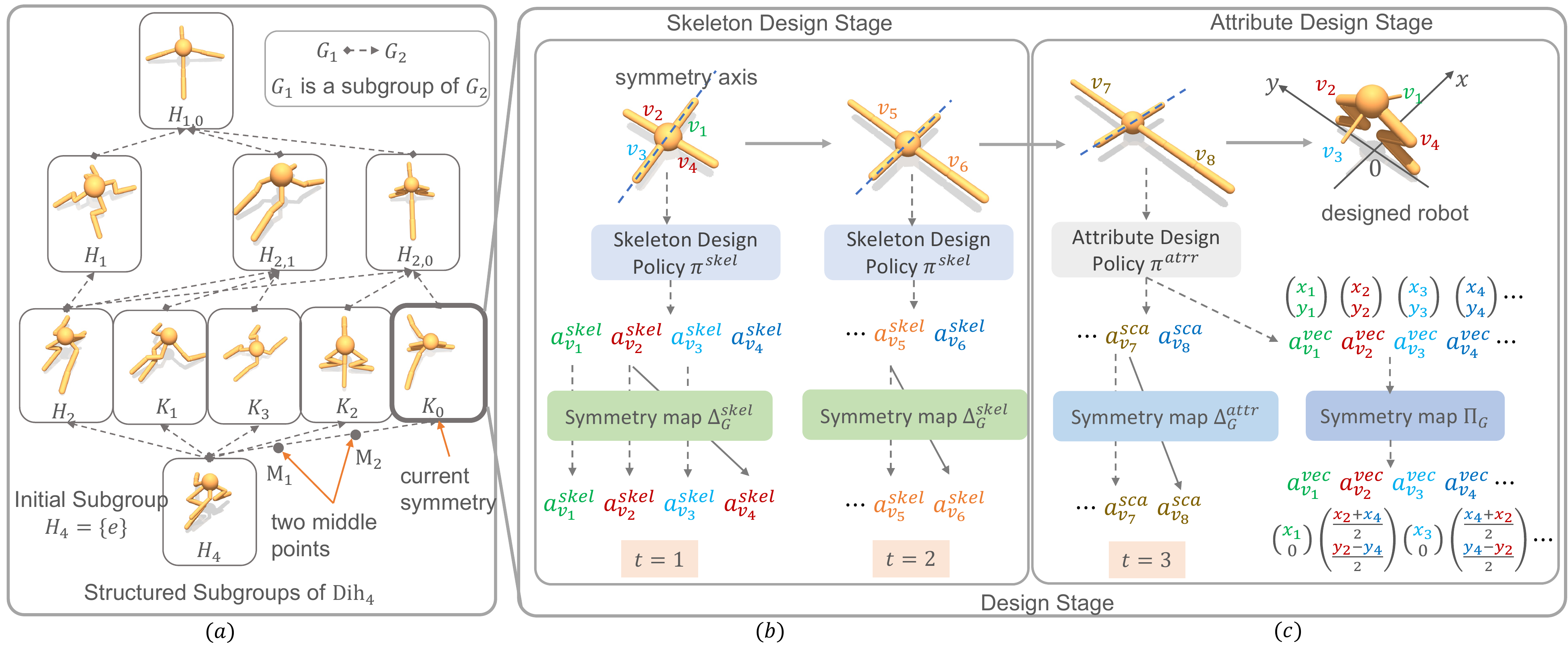}
\vspace{-2em}
\caption{Symmetry-aware robot design framework. (a) Search for the optimal symmetry in structured subgroups; (b) design the skeletal structures of the robot; (c) design the attributes of joints and limbs of the robot. Joints in the same color are in the same orbit in (b-c).}\label{fig:framework}
\vspace{-1em}
\end{figure*}

\textbf{Modular RL.} Robot design problem typically requires controlling robots with changeable morphologies, where the state and action spaces are incompatible across robots. This new issue cannot be tackled by traditional monolithic policies used in single-agent RL, but fortunately, modular RL that uses a shared policy to control each actuator separately holds the promise to solve it. Most prior works in this field represent the robot's morphology as a graph and use GNNs or message-passing networks as policies~\cite{wang2018nervenet,pathak2019assembling,huang2020one}. All of these GNN-like works show the benefits of modular policies over a monolithic policy in tasks tackling different morphologies. Recent works also proposed to use Transformer~\cite{vaswani2017attention} to represent policies to overcome the difficulty of message-passing in complex morphologies to further improve performance~\cite{kurin2020morphenus, gupta2021metamorph, dong2022solar}. In this paper, we build our method based on GNN-like policies for a fair comparison with the previous state-of-the-art baseline \tact{}~\cite{yuan2021transform2act} but note that our method is a plug-and-play module that can be used to any modular RL policies.

\textbf{Robot Design.} Automatic robot design problem aims at searching for robots that can be easily controlled and can perform various tasks efficiently. A line of works in this field focused only on designing the attributes of robots while ignoring skeletal structures~\cite{ha2017joint,yu2018policy,ha2019reinforcement}, which is limited in designing robot morphology. Another line of work considers both attribute design and skeleton design, which is known as combinatorial design optimization. Previous works mainly utilize evolutionary search (ES) algorithms for combinatorial design optimization~\cite{sims1994evolving,nolfi2000evolutionary,auerbach2014environmental,cheney2018scalable,jelisavcic2019lamarckian,zhao2020robogrammar}, which require robots with different structures to perform certain task independently and do not share experiences between robots. This will result in severe sample inefficiency and may require thousands of CPUs to finish one experiment~\cite{gupta2021embodied}. Recently, \citet{yuan2021transform2act} discussed this issue and used RL to optimize robot designs by incorporating design procedure into the decision-making process and formulating design optimization as learning a conditional policy. This method has shown great sample efficiency. However, none of the above studies have explored the structure of the design space and searched for robots directly from the vast design space, which may end up with abnormal robots with poor performance. As for works that considered symmetry in robot design, \nge{} \cite{wang2019neural} only considered bilateral symmetry, while our work considers a wide range of symmetries for various tasks.

\textbf{Symmetry in Real-World Creatures.} Our method that utilizes symmetry as the classification characteristic shares a similar and intriguing principle in real-world creatures, which might have been developed millions of years ago according to fossil evidence~\cite{evans2020discovery}. Certain symmetries are maintained by natural selection pressures and deviation from perfect symmetry is negatively correlated with species fitness~\cite{enquist1994symmetry}. \citet{raibert1986running} showed that during a series of bouncing and ballistic motions, symmetry contributes to achieving more complicated running behaviors, \eg{}, reciprocating leg symmetry is essential to make a quadruped gallop, symmetry of wings can help reduce energy expenditure in flight~\cite{polak1994science}. Our experiments in \cref{sec:exp} also showed similar conclusions that symmetry can help reduce control costs.



\section{Preliminaries}\label{sec:pre}

In this section, we introduce some background knowledge and notations necessary to present our method.


\textbf{Problem Settings.} We aim to search for a robot design $D$ from a design space $\gD$ to finish different tasks efficiently. A design $D$ includes the robot's skeletal structure, limb-specific attributes, and joint-specific attributions (\eg{}, limb length, size, and motor strength). Formally, a robot design can be represented by a graph $D=(V,E,Z)$, where each node $v\in V$ represents a joint in robot morphology and each edge $e=(v_i,v_j)\in E$ represents a limb connecting two joints $v_i,v_j$. Each $z\in Z$ is the attributes of a joint and the limb attached to this joint, including scalar values and vector values, and $|V|=|Z|$.


The designed robot then learns control policies to finish tasks by reinforcement learning (RL) algorithms. RL formulates a control problem as an infinite-horizon discounted Markov Decision Process (MDP), which is defined by a tuple $\gM=(\gS,\gA,\gT,\rho_0,R,\gamma)$, where these items represent state set, action set, transition dynamics, initial state distribution, reward function and discounted factor, respectively. For a fixed robot design and task, the objective of RL is to learn a control policy $\pi^C$ that maximizes the expected total discounted reward: $J(\pi^C)=\E_{\pi^C}[\sum_t \gamma^t r_t]$. 


Given that the robot design is changeable during learning, we condition the original transitions dynamics and reward function in $\gM$ on design $D$, yielding a more general transition dynamics $\gT(s_{t+1}|s_t,a_t,D)$ and reward function $R(s_t,a_t,D)$. Hence the reinforcement learning objective is also conditioned on $D$, and we optimize $J(\pi^C,D)$. Now the design optimization problem can be naturally formulated as a bi-level optimization problem~\cite{sinha2017reviewbilevel,colson2007overviewbilevel}: $D^* = \argmax_{D\in\gD} J(\pi^{C,*},D) \text{ s.t. } \pi^{C,*}=\argmax_{\pi^C} J(\pi^C, D)$
where the inner optimization problem is typically solved by RL but is especially computationally expensive for changeable robot designs. The outer optimization problem can be solved by evolutionary algorithms~\cite{wang2019neural, gupta2021embodied, sims1994evolving} or RL~\cite{yuan2021transform2act}. In this paper, we follow \tact{} to use RL as the outer problem solver for its efficiency compared with evolutionary algorithms.

\textbf{Group Theory.} We use group theory to represent different types of symmetries. Here we briefly introduce some notations and please refer to \cref{appx:group_theory} for details. A \textit{group} $G$ is a set with a binary operation such that it has four basic properties, \ie{}, associativity, closure, the existence of an identity ($e\in G$), and the existence of an inverse of each element ($g^{-1}\in G, \forall g\in G$)~\cite{gallian2021contemporary}. If a subset $H$ of $G$ is also a group under the same operation of $G$, we refer to it as a \textit{subgroup} of $G$ and denote it by $H<G$. 

A \textit{group action} is a function $G\times X\to X$ of group $G$ on some space $X$, and it satisfies $ex=x$, $(g h)x=g(hx), \forall g,h\in G, x\in X$. For an element $g\in G$, we define a \textit{transformation function} $\alpha_g^X: X\to X$ given by $x\mapsto gx$, which can be interpreted as a transformation of the point $x$ under group element $g$. For example, if $x$ is a robot, $g$ could be a rotation transformation of the robot along the z-axis passing through the torso of the robot. The \textit{orbit} of a point $x\in X$ is the set of all its transformation under $G$, denoted by $\gO_G(x)=\{\alpha_g^X(x)|\forall g\in G\}$. An important property is that $X$ can be partitioned by orbits: $X=\bigcup_{x\in X}\gO_G(x)$.

\textbf{Dihedral Group.} The dihedral group is a finite discrete group containing rotation and reflection transformations. A dihedral group $\text{Dih}_n (n\ge 3)$ can be generated by rotation transformation $\rho$ (counterclockwise rotation by $360^\circ/n$) and reflection transformation $\pi$ (reflection along x-axis). Concretely, $\text{Dih}_n\shorte \{\rho_k, \pi_{k-1}|k\shorte 1,2,\cdots,n\}$, where $\rho_k\shorte \rho^k, \rho_0\shorte \rho_n\shorte e$ and $\pi_k\shorte \rho^k\pi$. 
Each group element $g$ has multiple representations. We consider permutation representation $P_g$ and matrix representation $M_g$ in this paper. Considering the designed robot in \cref{fig:framework}(c) and taking $\pi_0\in \text{Dih}_4$ as an example, $P_{\pi_0}$ exchanges joint $v_2,v_4$ and $M_{\pi_0}$ reflects their coordinates along x-axis.

In this paper, we use the subgroups of the dihedral group to represent various symmetries. The subgroups of dihedral groups have three types: (1) $H_d\shorte \langle \rho_d \rangle$, where $1$$\le$$d$$\le$$n$ and $d|n$ ($n$ is divisible by $d$); (2) $K_i\shorte \langle \pi_i\rangle$, where $0$$\le$$i$$\le$$ n\shortn 1$; and (3) $H_{k,l}\shorte \langle\rho_k, \pi_l\rangle$, where $0$$\le$$l$$<$$k$$\le$$n\shortn 1$ and $k|n$. The group structure of $\text{Dih}_4$ is shown in \cref{fig:framework}(a). For more details, please refer to \cref{appx:dihedral_group}.

\section{Method}\label{sec:method}

In this section, we present our Symmetry-Aware Robot Design (\name{}) scheme that utilizes symmetry as a characteristic to exploit the structure of the design space and reduce the learning complexity. As in previous works on robot design, our learning framework consists of design searching and control policy learning. The focus of this paper is to incorporate symmetry into design search. To this end, our method consists of two major components: (1) searching for the optimal symmetry $G$ (\cref{sec:search_symmetry}) and (2) learning robot design  under the given symmetry $G$ (\cref{sec:design_a_robot}).



\subsection{Learning Robot Design under a Given Symmetry}\label{sec:design_a_robot}

\begin{algorithm}[htb]
   \caption{\name{}: Symmetry-Aware Robot Design}
   \label{alg:SARD_brief}
\begin{algorithmic}[1]
\INPUT {group $\text{Dih}_n$; number of intervals between symmetries $K$; symmetry sampling exploration rate $\epsilon$;}
\OUTPUT symmetry $G$; design policy $\pi^D$, control policy $\pi^C$; 
\STATE Initialize $\pi^D$ and $\pi^C$;
\STATE \CHANGE{(Sec.4.2):} Initialize symmetry $G\gets \{e\}$, value dict for symmetries $V\gets \mathbf{0}$;
\WHILE{not reaching max iterations}
\STATE Memory $\gM \gets \emptyset$;
\WHILE{$\gM$ not reaching batch size}
\STATE $D\gets$ initial robot design;

\WHILE{in Design Stage}
\STATE Sample design actions from $\pi^D$;
\STATE \CHANGE{(Sec.~4.1):} Transform design actions with symmetry maps $\Delta_G^{skel},\Delta_G^{atrr},\Pi_G$ (Eq.~(1) to (3));
\STATE Apply design actions to modify design $D$ and store them to $\gM$;
\ENDWHILE

\STATE Use $\pi^C$ to control current robot design $D$ and store trajectories to $\gM$.

\ENDWHILE
\STATE Update $\pi^C,\pi^D$ with PPO using samples in $\gM$;
\STATE \CHANGE{(Sec.4.2):} Update $V(G)\gets$ mean episode rewards in $\gM$ and sample a new symmetry from neighbors $G\gets \textit{Neighbor}(G)$ with $\epsilon$-greedy using $V$;
\ENDWHILE
\end{algorithmic}
\end{algorithm}

We now describe how to design robots that satisfy a given symmetry $G$. Symmetry refers to an object that is invariant under some transformations, and every subgroup of the dihedral group $\text{Dih}_n$ represents a type of symmetry, \eg{}, the designed robot in \cref{fig:framework}(c) is invariant under the symmetry represented by $K_0$ which contains a reflection transformation along the x-axis and the identity transformation. As the trivial subgroup that contains only identity transformation is considered, our method also covers the original design space. To formally represent symmetries, we define a $G$-symmetric property as follows:
\begin{definition}\label{def:G-symmetric}
A robot design $D=(V,E,Z)$ is $G$-symmetric if the robot is invariant under the transformation of group $G$.  Specifically, $\forall g\in G$, we have $D_g=D$, where $D_g\triangleq (V_g, E_g, Z_g)$,  $V_g\triangleq \{\alpha_g^V(v)|v\in V\}, E_g\triangleq \{\alpha_g^E(e)|e\in E\}, Z_g\triangleq \{\alpha_g^Z(z)|z\in Z\}$.
\end{definition}
where $\alpha_g^V,\alpha_g^E,\alpha_g^Z$ transform the design with group element $g$ and formal definitions are given in \cref{appx:def_transformation}. All $G$-symmetric robots constitute the $G$-symmetric space. Using the transformation function $\alpha_g^V(v)$, we can define the orbit of $v$: $\gO_G(v)=\{\alpha_g^V(v)|\forall g\in G\}$. For the designed robot in \cref{fig:framework}(c), joints $v_2, v_4$ are in the same orbit, \ie{}, $\gO_{K_0}(v_2)=\gO_{K_0}(v_4)=\{v_2,v_4\}$.

Most RL- or ES-based methods in robot design can be roughly divided into two stages: (1) \textbf{Design Stage}, where a new robot with design $D$ is generated from an initial design $D_0$ by a design policy $\pi^D$ or mutated by a random design policy $\pi^D$; (2) \textbf{Control Stage}, where the generated robot interacts with the environment using control policy $\pi^C$.

The \textbf{Design Stage} can be further divided into two sub-stages: \textit{\textbf{Skeleton Design Stage}} and \textit{\textbf{Attribute Design Stage}}, which generate the skeletal graph $(V,E)$ and attributes $Z$ of the design $D$, respectively. Thus the design policy $\pi^D$ will also be separated into two sub-policies: skeleton design policy $\pi^{skel}$ and attribute design policy $\pi^{attr}$, \ie{}, $\pi^D\shorte(\pi^{skel},\pi^{attr})$.

Every episode starts with the \textit{\textbf{Skeleton Design Stage}} by initializing a $G$-symmetric (where $G\shortl \text{Dih}_n$) initial design: $D_0\shorte (V\shorte \{v_1,v_2,\cdots,v_n\}, E\shorte \emptyset,Z\shorte \mathbf{0})$, \ie{}, the first robot in \cref{fig:framework}(b). In each time step, each joint $v\in V$ selects a discrete skeleton design action $a^{skel}_v\in\gA^{skel}$ based on the skeleton design policy $\pi^{skel}(a^{skel}_v|D, G)$. Here $D$ is the current design, $G$ is the given symmetry type and $\gA^{skel}$ is the skeleton action set including three actions: 1) AddJoint: joint $v$ will add a child joint $u$ to the skeletal graph; 2) DelJoint: joint $v$ will remove itself from the skeletal graph if it has no child joints; 3) NoChange: no changes will be made to joint $v$. All joints share the same action space. Then the robot design will transit to $D'$ and this sub-stage will last for $N^{skel}$ steps. To ensure that the designed robot is $G$-symmetric, we propose to keep the robot symmetric in each time step by making joints in the same orbit choose the same skeleton action, \eg{}, they could all choose the action selected by the joint with the smallest index. Therefore, joints $v_2,v_4$ in \cref{fig:framework}(b) will both use the action selected by $v_2$. This can be realized by a symmetry map $\Delta_G^{skel}:\gA^{skel}\to \gA^{skel}$, defined as follows:
\begin{equation}\label{equ:sym_skel}
    \Delta_G^{skel}(a^{skel}_v) = a_{\mu(\gO_G(v))}^{skel},\quad \forall v\in V,
\end{equation}
where $\gO_G(v)$ is $v$'s orbit under subgroup $G$, and $\mu: \gP(V)\to V$ returns the element with the smallest index in $\gO_G(v)$. $\gP(V)$ is the power set of $V$. We use $\{\Delta_G^{skel}(a^{skel}_v)\}_{v\in V}$ to generate a robot. In this way,  the new design $D'$ is also $G$-symmetric. 

At the \textit{\textbf{Attribute Design Stage}}, each joint $v\in V$ chooses a continuous attribute action $a^{attr}_v\in\gA^{attr}$ for $v$ and the limb attached to $v$ based on attribute design policy $\pi^{attr}(a^{attr}_v|D,G)$. Here $\gA^{attr}\shorte\gA^{sca}\times\gA^{vec}$ is the attribute action set containing scalar values $a^{sca}_v\in \gA^{sca}$ (motor strength, limb size, \etc{}) and vector values $a^{vec}_v\in \gA^{vec}$ (limb offset, \etc{}). Then the robot design will transit to $D'$ and this sub-stage will last for $N^{attr}$ steps. Like the first stage, we keep the robot symmetric in each time step to ensure that $D'$ is $G$-symmetric. For the scalar values, we define a similar symmetry map like \cref{equ:sym_skel}, $\Delta_G^{attr}:\gA^{sca}\to \gA^{sca}$:
\begin{equation}\label{equ:sym_attr}
    \Delta_G^{attr}(a^{sca}_v) = a_{\mu(\gO_G(v))}^{sca},\quad \forall v\in V.
\end{equation}
Therefore, joints $v_7,v_8$ in \cref{fig:framework}(c) will both adopt the action selected by $v_7$.

As for the vector values, we propose a novel symmetry map to project them to $G$-symmetric space. For simplicity, here we assume $a^{vec}_v=(x,y)^\top$, where $ x,y\in\sR$, only includes one coordinate here and $z$-value can be learned or set to a default value. The vector value actions of all joints form a matrix: $c=(a^{vec}_1, a^{vec}_2, \cdots, a^{vec}_{|V|})$. The coordinate $c$ should be invariant under transformation $g\in G$. Directly solving this problem is challenging, and we propose a novel symmetry map $\Pi_G: \sR^{2\times |V|}\to \sR^{2\times |V|}$ that can project any $c$ into $G$-symmetric space:
\begin{equation}\label{equ:symmetrizer}
    \Pi_G(c)=\frac{1}{|G|}\sum_{g\in G} M_gcP_{g^{-1}}.
\end{equation}
where $M_g$ and $P_g$ are the matrix and permutation representations of $g$. This property is verified in this theorem:
\begin{theorem}\label{thm:symmetrizer}
    The projected vector values $\Pi_G(c)$ defined in \cref{equ:symmetrizer} are $G$-symmetric and, if $c$ is already $G$-symmetric, then $\Pi_G(c)=c$.
\end{theorem}
This theorem implies two facts: (1) realizability: the projected vector values are $G$-symmetric; and (2) completeness: $\Pi_G(c)$ can cover the whole $G$-symmetric space. The proof is provided in \cref{appx:symmetrizer} and \cref{fig:framework}(c) shows the transformation results of $K_0$, which is also derived in \cref{appx:symmetrizer}. It is straightforward to extend one coordinate to multiple coordinates by applying \cref{equ:symmetrizer} coordinate-wise. Thus, we can use $\Pi_G(c)$ as the vector value of attribute design actions to ensure that $D'$ is $G$-symmetric. Putting all symmetry maps together, we prove in \cref{appx:symmetry_all} the following theorem:
\begin{theorem}\label{thm:symmetry}
    The transformed design space by transformations $\Delta_G^{skel}$ (\cref{equ:sym_skel}), $\Delta_G^{attr}$ (\cref{equ:sym_attr}) and $\Pi_G$ (\cref{equ:symmetrizer})  is equivalent to $G$-symmetric space.
\end{theorem}
In the Design Stage, robots do not interact with the environments and will not receive rewards. The design policy is trained using PPO with rewards from the Control Stage. 

The \textbf{Control Stage} is the same as the normal robot control problem, except that the control policy $\pi^C(a|s,D)$ is conditioning on the current design. \cref{fig:control}(c) shows a trajectory of the designed robots of \cref{fig:framework}(c) in the Control Stage for \locomotionft{} task. To incorporate $D$ in to policies we implement policies $\pi^D$ and $\pi^C$ with graph neural networks~\cite{scarselli2008graph, bruna2013spectral, kipf2016semi}, which are optimized by Proximal Policy Optimization (PPO)~\cite{schulman2017PPO}, a standard policy gradient method~\cite{williams1992PG}.

\begin{figure*}[htb]
\centering
\includegraphics[width=0.9\linewidth]{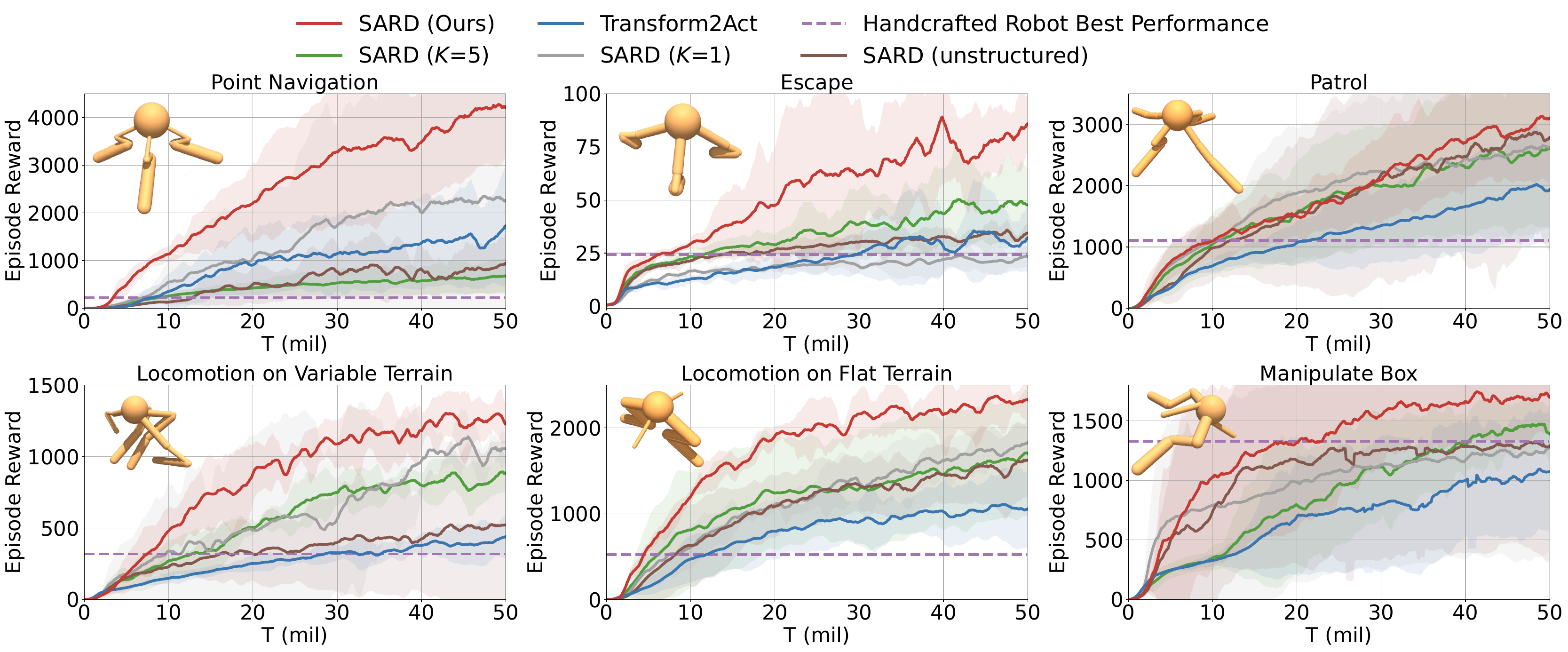}
\vspace{-1em}
\caption{Training performance of \name{} compared against baselines and ablations}\label{fig:training}
\vspace{-1em}
\end{figure*}

\begin{figure*}[htb]
\centering
\includegraphics[width=0.9\linewidth]{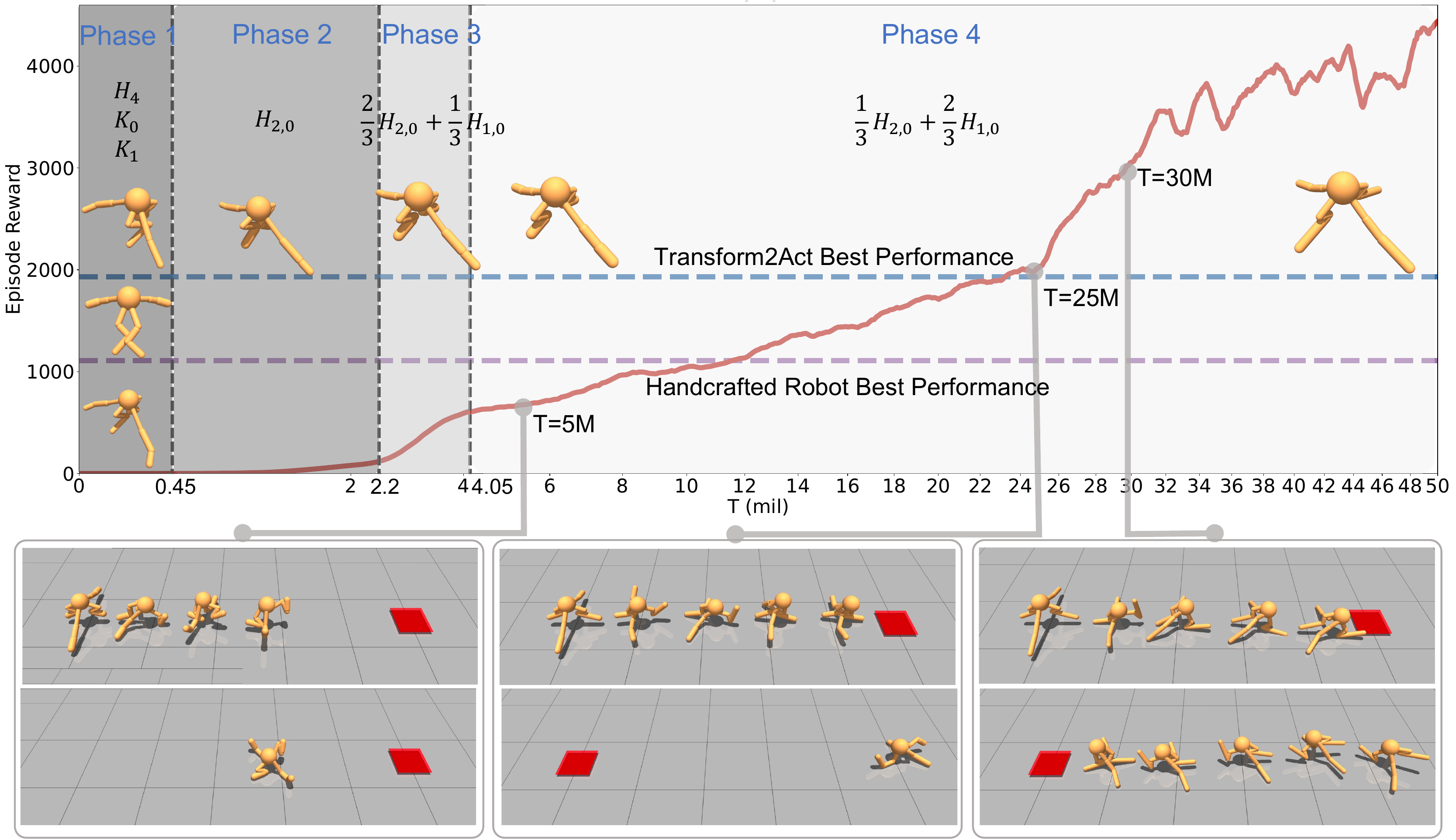}
\vspace{-1em}
\caption{Visualization of the learning process of \name{}. x-axis has been rescaled to better display early phases.}\label{fig:evolution}
\vspace{-1em}
\end{figure*}

\subsection{Searching for the Optimal Symmetry}\label{sec:search_symmetry}

We now discuss how to find the optimal symmetry for differing tasks. One simplest way is to sample several subgroups from $\text{Dih}_n$, evaluate them, and choose the subgroup with the highest performance every iteration. However, in this way, the sampled symmetry type in consecutive iterations $G_i$ and $G_{i+1}$ might be dissimilar and thus the designed robots under these two symmetries might be vastly different, which may cause gradient conflict problems~~\cite{liu2021conflict,javaloy2021rotograd,shi2021gradient} as the control policy $\pi^C(a|s,D)$ is shared across all robots.


To mitigate this problem and smooth the gap between subgroups, we propose a novel search method by exploiting the structure of subgroups as in \cref{fig:framework}(a). The core idea is to let subgroups $G_i$ and $G_{i+1}$ in consecutive iterations be similar by ensuring that they are adjacent in group structure: $G_{i+1}\in \textit{Neighbor}(G_{i})$, which is defined in \cref{appx:subgroup_neighbor}. In \cref{fig:framework}, $\textit{Neighbor}(K_0)=\{H_4,K_0,H_{2,0}\}$.

In practice, we maintain a value dict for all subgroups based on mean episode reward. At the beginning of training, we set initial subgroup $G_0=\{e\}$ which represents the original design space to avoid introducing any prior knowledge, and then in each iteration $i\shortp 1$, we sample a subgroup from $\textit{Neighbor}(G_{i})$ using $\epsilon$-greedy based on their values. The sampled subgroup $G_{i+1}$ is then used to generate robots in \cref{sec:design_a_robot} for several episodes. At the end of each iteration, we update the value for the current symmetry. 

However, it is possible that $G_i, G_{i+1}$ are not similar enough. See \cref{fig:framework}(a) for intuition. To further smooth the gap between subgroups, we consider the middle points between two adjacent subgroups $G,G'$. Assuming $G<G'$, there is no subgroup between them as we discussed above. However, we only need to ensure that these middle points can be used in Skeleton Design Stage (\cref{equ:sym_skel}) and Attribution Design Stage (\cref{equ:sym_attr,equ:symmetrizer}). We can prove that $\Pi_{G'}(c) = \beta_0 \Pi_{G}(c) + (1-\beta_0)\Pi_{G'\shortn G}(c)$
where $\beta_0=|G|/|G'|$ and $G'\shortn G\triangleq \{g|g\in G', g\notin G\}$ (see \cref{appx:derive_beta}). We therefore define
\begin{align}\label{equ:symmetrizer_beta}
    \Pi_{G,G',\beta}(c) \triangleq \beta\Pi_{G}(c) + (1-\beta)\Pi_{G'\shortn G}(c)
\end{align}
where for any $\beta$ in interval $\left[\beta_0,1\right]$, $\Pi_{G,G',\beta}$ is the symmetry map of a middle point between $G,G'$. $\Pi_{G,G',\beta_0}\shorte\Pi_{G'}$ and $\Pi_{G,G',1}\shorte\Pi_{G}$. In practice, we divide the interval equally into $K$ parts and consider the middle points as neighbors. For example, in \cref{fig:framework}(a), $K\shorte 3$, denoting $M_2$ by $\frac{1}{3}H_4+\frac{2}{3}K_0$, we have $\textit{Neighbor}(K_0)=\{\frac{1}{3}H_4+\frac{2}{3}K_0, \frac{1}{3}H_{2,0}+\frac{2}{3}K_0,K_0\}$. 
In this way, the designed robots in consecutive iterations are much more similar than before. And we can prove a similar theorem as \cref{thm:symmetrizer} in \cref{appx:symmetrizer_beta}. 

We outline our SARD algorithm in \cref{alg:SARD_brief} and we also provide a detailed version in \cref{appx:algo}.


\section{Experiments}\label{sec:exp}

In this section, we benchmark our method \name{} on various MuJoCo~\cite{todorov2012mujoco} tasks. We evaluate the effectiveness of \name{} by asking the following questions: (1) Can robot design help improve performance compared with handcrafted design? (\cref{sec:training}) (2) Can \name{} outperform other robot design baselines in various tasks? (\cref{sec:training}) (3) How does \name{} search for the desired symmetry and how does symmetry facilitate control policy learning? (\cref{sec:evolution}) (4) Can the searched symmetry by \name{} generalize to all tasks? (\cref{sec:fix_sym}). For qualitative results, please refer to the videos on our anonymous project website\footnote[1]{\url{https://sites.google.com/view/robot-design}}. And our code is available at GitHub\footnote[2]{\url{https://github.com/drdh/SARD}}.

\subsection{Experiment Setup}
We run experiments on six tasks in this section. All runs are conducted with 4 random seeds and the mean performance as well as 95\% confidence intervals are shown.

\textbf{Environments.} All six tasks are adapted from \citet{gupta2021embodied}, which are created based on MuJoCo~\cite{todorov2012mujoco} physics simulator. The tasks vary in objectives (locomotion, approaching random goals, pushing an object,\etc{}), terrains (flat terrain, variable terrain, bowl-shaped terrain,\etc{}), and observation space. Here we briefly discuss the tasks: (1) \pointnav{}. An agent is spawned at the center of a flat arena and has to reach a random goal in the arena; (2) \escape{}. An agent is spawned as the center of a bowl-shaped terrain and has to escape from the region; (3) \patrol{}. An agent needs to run forth and back between two goals; (4) \locomotionvt{}. An agent has to run forward as fast as possible in a variable terrain; (5) \locomotionft{}. An agent has to run forward as fast as possible in flat terrain; (6) \manipulationbox{}. An agent is tasked with pushing a box into a randomly generated goal. For more details, please refer to \cref{appx:task_detail}.

\textbf{Baselines and Implementations.} Our method \name{} is implemented on the top of \tact{}~\cite{yuan2021transform2act}, which is the previous state-of-the-art robot design method and is compared in this section. We use the same hyperparameters as \tact{} for fair comparisons. To show the strength of robot design, we also compare \name{} with \originalant{}, which is the human-designed Ant using expert knowledge from OpenAI. We did not include any ES-based methods here because their sample inefficiency has been verified by \citet{yuan2021transform2act}.

\textbf{Ablations.} There are two contributions that characterize our method. (1) A plug-and-play transformation module that is used to generate robots under a given symmetry. (2) A search method that utilizes structured subgroups to find the optimal symmetry. Our novelties are mainly about the searching and utilization of symmetries, and these two contributions closely rely on each other. If we remove symmetry, we will get \tact{}. Therefore, the effectiveness of symmetry-aware robot design can be demonstrated by comparing \tact{} and \name{}. To validate the effectiveness of the searching method, we further design the following ablation studies: (1) \nameUnstruct{}. Do not consider the structure of subgroups discussed in \cref{sec:search_symmetry} and sample subgroups directly from the whole set. (2) \nameKone{} and \nameKfive{}. Our method divides the interval between two adjacent subgroups into $K$($\shorte3$ in \name{}) parts. We use these two ablations to show the effectiveness of smoothing the gap between subgroups. 

\begin{figure*}[ht]
\centering
\includegraphics[width=0.9\linewidth]{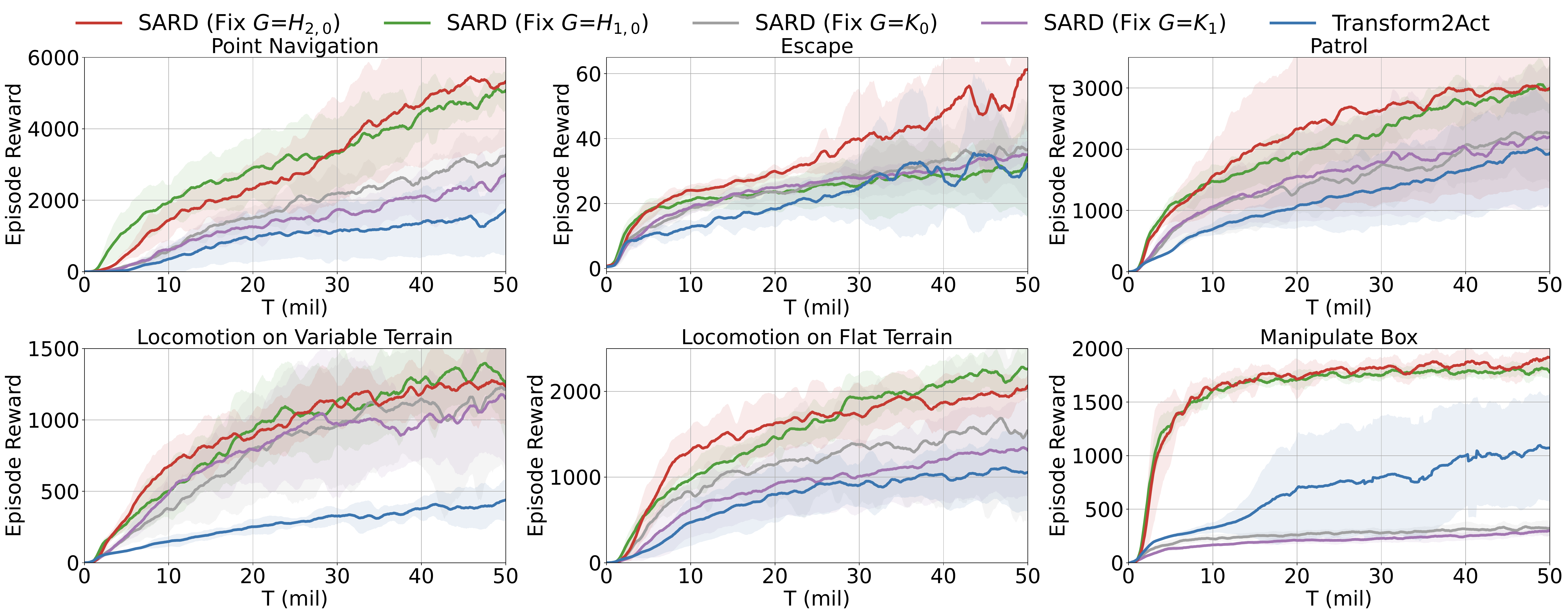}
\vspace{-1em}
\caption{Generalization performance of the symmetry searched by \name{} against baselines.}\label{fig:fix_sym}
\vspace{-1em}
\end{figure*}

\subsection{Training Performance Comparison}\label{sec:training}
We summarize the training performance in \cref{fig:training} and we also show one representative robot designed by \name{} at the end of training in the upper left corner of each sub-figure to provide more intuition. \name{} outperforms the previous state-of-the-art algorithm \tact{} and handcrafted robot Ant in all tasks. This validates that (1) the designed robots can improve final performance compared with handcrafted robots and (2) symmetry-aware learning can effectively help solve the robot design problem. 

As for ablations, \name{} is better than \nameUnstruct{} in all tasks, which shows the usefulness of utilizing subgroup structures. Further, \name{} outperforms \nameKone{} in most tasks, which indicates the effectiveness of smoothing the gap between consecutive subgroups. Interestingly, \name{} also outperforms \nameKfive{}. This is because dividing the interval between two adjacent subgroups into too many parts will slower the learning such that it will take more iterations to move from one symmetry to another and the training could easily get stuck.

\subsection{Robot Design Analysis}\label{sec:evolution}
To investigate why \name{} performs better than baselines and how can symmetry help improve performance, in this subsection, we visualize the learning process of \name{} in \patrol{} task in \cref{fig:evolution}. According to the change of symmetry, the learning process can be divided into four phases. And in each phase, we also show the corresponding symmetry and a representative robot of this phase. In Phase 1, \name{} explores some symmetries around the initial subgroup $H_4$ (see \cref{fig:framework}(a)) and changed symmetries frequently. In Phase 2, \name{} chooses a new subgroup $H_{2,0}$ and the episode reward is still low. In Phase 3, \name{} moves to a middle point between two subgroups $H_{2,0}$ and $H_{1,0}$ that is closer to $H_{2,0}$ and the learning curve starts to rise, which indicates the usefulness of the smoothing trick of \name{}. In Phase 4, \name{} finally finds a suitable subgroup that still lies between two subgroups but closer to $H_{1,0}$. This subgroup is then left unchanged and \name{} optimizes robot designs under this symmetry, in which case the search space is legitimately reduced. The left and right robots in Phase 4 are loaded from the beginning and end of this phase, respectively. Only some attributes of the robots are changed. 

We also visualize the time-lapse images of control policies in \cref{fig:evolution} to show how symmetry can foster learning to control. \patrol{} task requires robots to run forth and back between two goals as fast as possible. At $t\shorte 5M$, \name{} still has trouble in running to the goal ahead. At $t\shorte 25M$, \name{} learns to reach the goal ahead but fails to learn to run backward and \tact{} also gets stuck at this point. However, \name{} eventually exceeds the best performance of \tact{} at $t\shorte 30M$. This is because, in \patrol{} task, \tact{} usually needs to learn to turn around, which can easily lead to a tumble and is hard for robots. But for \name{}, thanks to its symmetry, it does not have to turn around and can directly run in the opposite direction, which shows one of the benefits of symmetry. 

The visualization results reflect two facts: (1) symmetry in robots is helpful in terms of reducing the control complexity, and (2) \name{} can help learn symmetry by the means of searching in structured subgroups.

\subsection{Generalization of the Learned Symmetry}\label{sec:fix_sym}
To further validate the generalization of the learned symmetry by \name{}. We counted all results of \name{} in \cref{fig:training} and if the symmetry is a middle point between two subgroups, we choose to count the one closest to it. We find that $3/4$ of the experiments ended up with $H_{2,0}$ and $H_{1,0}$ ($H_{2,0}$ accounted for 45.83\% and $H_{1,0}$ accounted for 29.17\%). Therefore, in this subsection, we run new experiments that fix symmetry to several potential ones to further test the effectiveness of \name{}. The results are summarized in \cref{fig:fix_sym}. Here $K_1$ and $K_0$ is the bilateral symmetry used in previous works~\cite{gupta2021embodied, wang2019neural}. \name{} (Fix $G=H_{2,0}$) and \name{} (Fix $G=H_{1,0}$) outperform other symmetries in most tasks, which validate the generalization of the learned symmetries.

\section{Conclusions}
In this paper, we exploit the structure of design space in the robot design problem with the symmetry characteristics of robot. Our method \name{} can generate efficient symmetric robots while still covering the original design space. The theoretical analyses and empirical evaluations of \name{} have shown its superior strength.

\section*{Acknowledgments}
This work is supported in part by Science and Technology Innovation 2030 - ``New Generation Artificial Intelligence'' Major Project (No. 2018AAA0100904) and the National Natural Science Foundation of China (62176135).


\bibliography{icml2023}
\bibliographystyle{icml2023}

\newpage
\appendix
\onecolumn

\section{Method Details}\label{appx:method}

\subsection{Algorithm}\label{appx:algo}
We implement \name{} based on \tact{} in \cref{sec:exp} and outline the algorithm in \cref{alg:SARD}. The highlighted lines are the changed procedures \tact{}. The main components of this algorithm are:

\textbf{Line 2, 29-30}: Initialize and update current symmetry type;

\textbf{Line 7-12}: Skeleton Design Stage, in which \name{} applies skeleton symmetry map $\Delta_G^{skel}$ (defined in \cref{equ:sym_skel}) to transform skeleton design actions.

\textbf{Line 13-19}: Attribute Design Stage, in which \name{} applies attribute symmetry map $\Delta_G^{attr}$ (defined in \cref{equ:sym_attr}) and $\Pi_G$ (defined in \cref{equ:symmetrizer_beta}) to transform attribute design actions.

\textbf{Line 20-26}: Control Stage.

\begin{algorithm}[htb]
   \caption{\name{}: Symmetry-Aware Robot Design implemented based on \tact{}}
   \label{alg:SARD}
\begin{algorithmic}[1]
\INPUT \CHANGE{group $\text{Dih}_n$; number of intervals between symmetries $K$; symmetry sampling exploration rate $\epsilon$;} skeleton design steps $N^{skel}$; attribute design steps $N^{attr}$; max steps of each episode $H$;
\OUTPUT symmetry $G$; design policy $\pi^D=(\pi^{skel},\pi^{attr})$, control policy $\pi^C$; 
\STATE Initialize $\pi^D$ and $\pi^C$; 
\STATE \CHANGE{Initialize symmetry $G\gets \{e\}$; initialize value dict for symmetries $V\gets \mathbf{0}$;}
\WHILE{not reaching max iterations}
\STATE Memory $\gM \gets \emptyset$;
\WHILE{$\gM$ not reaching batch size}
\STATE $D_0\gets$ initial robot design;

\FOR{$t=1,2,\ldots,N^{skel}$}
\STATE Sample skeleton design action $a_{v,t}^{skel} \sim \pi^{skel}$;
\STATE \CHANGE{Apply skeleton symmetry map $\Delta_G^{skel}$ defined in \cref{equ:sym_skel} to transform $a_{v,t}^{skel}$;}
\STATE $D_{t+1}\gets$ apply $\{a_{v,t}^{skel}\}_{v\in V_t}$ to modify skeleton $(V_t,E_t)$ in $D_t$;
\STATE $r_0\gets 0$; store $(r_t,\{a_{v,t}^{skel}\}_{v\in V_t}, D_t)$ into $\gM$;
\ENDFOR

\FOR{$t=N^{skel}\shortp 1,\ldots, N^{skel}\shortp N^{attr}$}
\STATE Sample attribute design action $(a_{v,t}^{sca},a_{v,t}^{vec})=a_{v,t}^{attr}\sim \pi^{attr}$;
\STATE \CHANGE{Apply attribute symmetry map for scalar values $\Delta_G^{atrr}$ defined in \cref{equ:sym_attr} to transform $a_{v,t}^{sca}$;}
\STATE \CHANGE{Apply attribute symmetry map for vector values $\Pi_G$ defined in \cref{equ:symmetrizer_beta} to transform $a_{v,t}^{vec}$;}
\STATE $D_{t+1}\gets$ apply $\{a_{v,t}^{attr}\}_{v\in V_t}$ to modify attributes $Z_t$ in $D_t$;
\STATE $r_t\gets 0$; store $(r_t, \{a_{v,t}^{attr}\}_{v\in V_t},D_t)$ into $\gM$;
\ENDFOR

\STATE $s_{t+1}\gets$ initial environment state;
\FOR{$t=N^{skel}\shortp N^{attr}\shortp 1,\ldots,H$}
\STATE Sample control actions $a^C_t\sim \pi^C$;
\STATE $s_{t+1}\gets$ environment dynamics $\gT(s_{t+1}|s_t,a^C_t,D_t)$; 
\STATE $r_t\gets$ environment reward $R(s_t,a^C_t,D_t)$; $D_{t+1}\gets D_t$;
\STATE Store $(r_t,a_t^C,s_t,D_t)$ into $\gM$;
\ENDFOR
\ENDWHILE
\STATE Update $\pi^C,\pi^D$ with PPO using samples in $\gM$;
\STATE \CHANGE{Update symmetry values $V(G)\gets$ mean episode rewards in $\gM$;}
\STATE \CHANGE{Sample a new symmetry from neighbors $G\gets \textit{Neighbor}(G)$ with $\epsilon$-greedy using symmetry values $V$;}
\ENDWHILE
\end{algorithmic}
\end{algorithm}

\subsection{Group Theory}\label{appx:group_theory}
A group $G$ is a set with a binary operation ($\cdot$) such that it has four basic mathematical properties~\cite{gallian2021contemporary}:
\begin{itemize}
    \item associativity: $g_1\cdot(g_2\cdot g_3)=(g_1\cdot g_2)\cdot g_3, \forall g_1,g_2,g_3 \in G$.
    \item closure: $g_1\cdot g_2 \in G, \forall g_1, g_2\in G$.
    \item the existence of an identity: $e\in G$
    \item the existence of an inverse of each element: $g^{-1}\in G, \forall g\in G$.
\end{itemize}

\subsection{Dihedral Groups}\label{appx:dihedral_group}
\textbf{Definition.} The dihedral group is a finite discrete group containing rotation and reflection transformation. A dihedral group $\text{Dih}_n (n\ge 3)$ can be generated by rotation transformation $\rho$ (counterclockwise rotation by $360^\circ/n$) and reflection transformation $\pi$ (reflection along x-axis), \ie{}, $\text{Dih}_n=\langle \rho, \pi | \rho^n=\pi^2=1, \pi\rho\pi^{-1}=\rho^{-1}\rangle$. Concretely, $\text{Dih}_n=\{\rho_k, \pi_{k-1}|k=1,2,\cdots,n\}$, where $\rho_k=\rho^k, \rho_0=\rho_n=e$ and $\pi_k=\rho^k\pi$. 

\textbf{Group Element Representation.} Each group element can have multiple representations. We consider permutation representation and matrix representation in this paper. The permutation and matrix representations of $g$($\in \text{Dih}_4$)) are list below:
\begin{align*}
    &P_{\rho_0} = \begin{pmatrix}
        1 & 0 & 0 & 0\\
        0 & 1 & 0 & 0\\
        0 & 0 & 1 & 0\\
        0 & 0 & 0 & 1
    \end{pmatrix},
    M_{\rho_0}= \begin{pmatrix}
        1 & 0 \\
        0 & 1
    \end{pmatrix},
    &P_{\rho_1} = \begin{pmatrix}
        0 & 0 & 0 & 1\\
        1 & 0 & 0 & 0\\
        0 & 1 & 0 & 0\\
        0 & 0 & 1 & 0
    \end{pmatrix},
    M_{\rho_1}= \begin{pmatrix}
        0 & -1 \\
        1 & 0
    \end{pmatrix}, \\
    &P_{\rho_2} = \begin{pmatrix}
        0 & 0 & 1 & 0\\
        0 & 0 & 0 & 1\\
        1 & 0 & 0 & 0\\
        0 & 1 & 0 & 0
    \end{pmatrix},
    M_{\rho_2}= \begin{pmatrix}
        -1 & 0 \\
        0 & -1
    \end{pmatrix}, 
    &P_{\rho_3} = \begin{pmatrix}
        0 & 0 & 0 & 1\\
        1 & 0 & 0 & 0\\
        0 & 1 & 0 & 0\\
        0 & 0 & 1 & 0
    \end{pmatrix},
    M_{\rho_3}= \begin{pmatrix}
        0 & 1 \\
        -1 & 0
    \end{pmatrix}, \\
    &P_{\pi_0} = \begin{pmatrix}
        1 & 0 & 0 & 0\\
        0 & 0 & 0 & 1\\
        0 & 0 & 1 & 0\\
        0 & 1 & 0 & 0
    \end{pmatrix},
    M_{\pi_0}= \begin{pmatrix}
        1 & 0 \\
        0 & -1
    \end{pmatrix}, 
    &P_{\pi_1} = \begin{pmatrix}
        0 & 1 & 0 & 0\\
        1 & 0 & 0 & 0\\
        0 & 0 & 0 & 1\\
        0 & 0 & 1 & 0
    \end{pmatrix},
    M_{\pi_1}= \begin{pmatrix}
        0 & 1 \\
        1 & 0
    \end{pmatrix}, \\
    &P_{\pi_2} = \begin{pmatrix}
        0 & 0 & 1 & 0\\
        0 & 1 & 0 & 0\\
        1 & 0 & 0 & 0\\
        0 & 0 & 0 & 1
    \end{pmatrix},
    M_{\pi_2}= \begin{pmatrix}
        -1 & 0 \\
        0 & 1
    \end{pmatrix}, 
    &P_{\pi_3} = \begin{pmatrix}
        0 & 0 & 0 & 1\\
        0 & 0 & 1 & 0\\
        0 & 1 & 0 & 0\\
        1 & 0 & 0 & 0
    \end{pmatrix},
    M_{\pi_3}= \begin{pmatrix}
        0 & -1 \\
        -1 & 0
    \end{pmatrix}, \\
\end{align*}
Here we take $\pi_0\in \text{Dih}_4$ as an example. $P_{\pi_0}$ exchanges $2,4$-columns and keeps the other unchanged when right multiplied with coordinate pairs ($\sR^{2\times 4}$), and $M_{\pi_0}$ reflects the coordinates along x-axis when left multiplied with the original coordinates. Taking the designed robot in \cref{fig:framework}(c) as an example, $P_{\pi_0}$ exchanges joint $v_2,v_4$ and $M_{\pi_0}$ reflects their coordinates along x-axis. 

\textbf{Subgroups.} The subgroups of dihedral groups can be classed into three categories: 
\begin{enumerate}
    \item $H_d=\langle \rho_d \rangle$, where $1\le d \le n$ and $d|n$ ($n$ is divisible by $d$).
    \item $K_i=\langle \pi_i\rangle$, where $0\le i\le n\shortn 1$.
    \item $H_{k,l}=\langle\rho_k, \pi_l\rangle$, where $0\le l < k \le n\shortn 1$ and $k|n$. 
\end{enumerate}
The group structure of $\text{Dih}_4$ is shown in \cref{fig:framework}(a). An interesting result is that the number of the subgroups of $\text{Dih}_n$ is $\sum_{w\in W}(1+w)$, where $W$ is the set of all positive divisors of $n$. For example, $W=\{1,2,4\}$ for $\text{Dih}_4$ and thus it has 10 subgroups. 

We list all subgroups of $\text{Dih}_4$ below:
\begin{align*}
    &H_1 = \{\rho_0,\rho_1,\rho_2,\rho_3\},\\
    &H_2 = \{\rho_0,\rho_2\},\\
    &H_4 = \{\rho_0\},\\
    &K_0 = \{\rho_0,\pi_0\},\\
    &K_1 = \{\rho_0,\pi_1\},\\
    &K_2 = \{\rho_0,\pi_2\},\\
    &K_3 = \{\rho_0,\pi_3\},\\
    &H_{1,0} = \{\rho_0,\rho_1,\rho_2,\rho_3,\pi_0,\pi_1,\pi_2,\pi_3\},\\
    &H_{2,0} = \{\rho_0,\rho_2,\pi_0,\pi_2\},\\
    &H_{2,1} = \{\rho_0,\rho_2,\pi_1,\pi_3\}.\\
\end{align*}

\textbf{Extend the Dimension of $P_g$.} The original permutation representation $P_{g}\in\{0,1\}^{n\times n}$, we extend it to $\{0,1\}^{|V|\times|V|}$. In the design $D=\{V,E,Z\}$, the number of joints near the torso is exactly $n$. And for other joints in layer $k$, we only need to use their parent joints' permutation representation. Specifically, we select the corresponding rows and columns of permutation representation belonging to their parent joints and denote it by $P_k$, then the new permutation is:
\begin{equation*}
    P_g = \begin{pmatrix}
        P_g &     &     &        &     \\
            & P_2 &     &        &     \\
            &     & P_3 &        &     \\
            &     &     & \ddots &     \\
            &     &     &        & P_J 
    \end{pmatrix}
\end{equation*}
where $J$ is the total number of layers (see \cref{appx:def_transformation} for details).

\subsection{Formal Definitions of the Transformation Functions}\label{appx:def_transformation}
We now define the transformation functions used in $G$-symmetric in \cref{def:G-symmetric}: $\alpha_g^V(v),\alpha_g^E(e),\alpha_g^Z(z)$.

Based on the distance to the root node (torso in robots), nodes $V$ can be divided into several disjoint subsets: $V=\bigcup_{j=1}^J V_j$, where the distance of $v\in V_j$ to root node is $j$. And thus the attribute set $Z$ can also be divided: $Z=\bigcup_{j=1}^J Z_j$. Note that each attribute corresponds to a joint, thus we always have $|V_j|=|Z_j|,\forall j$. Assuming the current dihedral group is $\text{Dih}_n$, to ensure the designed robot is $G$-symmetric ($G\shortl \text{Dih}_n$), we let $|V_1|=n$ by setting the initial design $D_0=(V=\{v_1,v_2,\cdots,v_n\}, E=\emptyset,Z=\mathbf{0})$ and forbidding growing any new joints from the root node (torso). We further set the embedding of $v_j$ to $\delta_j\in\{0,1\}^{1\times n}$, where only the $j$-th row of $\delta_j$ is $1$. And define $Enc(v_j)=\delta_j, Dec(\delta_j)=v_j$, we have
\begin{equation}
    \alpha_g^V(v) = Dec(Enc(v)^\top P_{g}), \forall v\in V_1,
\end{equation}
where $P_g$ is the permutation representation of $g$. For the nodes in other layers, the embeddings are the of their parent nodes and $\alpha_g^V$ the same. Using $\alpha_g^V$, we have:
\begin{equation}
    \alpha_g^E(e)=\alpha_g^E((v_i,v_j))=(\alpha_g^V(v_i),\alpha_g^V(v_j)),\forall e\in E.
\end{equation}
As for $\alpha_g^Z(z)$, where $z=(z^{sca},z^{vec})$, the scalar value $z^{sca}$ is unchanged and the vector value $z^{vec}$ is transformed using $M_g$, \ie{}, the matrix representation of $g$:
\begin{equation}
    \alpha_g^Z(z)=\alpha_g^Z((z^{sca},z^{vec}))=(z^{sca},M_gz^{vec}).
\end{equation}

\subsection{Proof of \cref{thm:symmetrizer} of $\Pi_G$}\label{appx:symmetrizer}
To prove \cref{thm:symmetrizer}, we first discuss the feasible symmetric space of $a^{vec}_v$, then prove the symmetry map can project any design into this symmetric space and leave any design in the symmetric space unchanged. For simplicity, here we assume $a^{vec}_v=(x,y)^\top$, where $ x,y\in\sR$, only includes one coordinate. The vector value actions of all joints form a matrix:
\begin{equation}
    c=(a^{vec}_1, a^{vec}_2, \cdots, a^{vec}_{|V|})
    =
    \begin{pmatrix}
    x_1 & x_2 & \cdots, x_{|V|} \\
    y_1 & y_2 & \cdots, y_{|V|} 
    \end{pmatrix}
\end{equation}
The coordinate $c$ should be invariant under transformation $g\in G$. That is, the transformed coordinate $c'\shorte M_{g}c$ should be the same as $c$, where $M_{g}$ is the matrix representation of $g$. In \cref{fig:framework}(c), we only consider the first four joints here and have
\begin{align}
    c'=M_{\pi_0}c = \begin{pmatrix}
        1 & 0 \\
        0 & -1
    \end{pmatrix}
    \begin{pmatrix}
        x_1 & x_2 & x_3 & x_4 \\
        y_1 & y_2 & y_3 & y_4
    \end{pmatrix} = \begin{pmatrix}
        x_1 & x_2 & x_3 & x_4 \\
        -y_1 & -y_2 & -y_3 & -y_4
    \end{pmatrix}.
\end{align}
However, the order of the items in $c$ and $c'$ might not be aligned, \eg{}, $v_2$ in $c'$ should be aligned with $v_4$ in $c$. To solve this issue, we inversely permutate the order of the items in $c'$ by left multiplying it with the permutation representation of $g^{-1}$: $P_{g^{-1}}$. 
 Since $\pi_0^{-1}=\pi_0$, we have
 \begin{equation}
    c' P_{\pi_0} = \begin{pmatrix}
        x_1 & x_4 & x_3 & x_2 \\
        -y_1 & -y_4 & -y_3 & -y_2
    \end{pmatrix}.
 \end{equation}
Note that the original permutation representation $P_{g^{-1}}\in\{0,1\}^{n\times n}$, we extend it to $\{0,1\}^{|V|\times|V|}$ in \cref{appx:dihedral_group}. 
Finally we need to ensure that $M_gcP_{g^{-1}}=c, \forall g\in G$. That is,
\begin{equation}\label{equ:coord_set}
    c\in \gC_G \triangleq \{c\in\sR^{2\times |V|} | M_gcP_{g^{-1}}=c,\forall g\in G\}.
\end{equation}
Now we prove that The symmetry map $\Pi_G$ defined in \cref{equ:symmetrizer} projects any coordinate $c\in \sR^{2\times |V|}$ into the set $\gC_G$ defined in \cref{equ:coord_set}. And if $c\in\gC_G$, the symmetry map will leave $c$ unchanged. Similar proofs are provided in \citet{van2020mdp} for a different purpose.

For $\forall h\in G, c\in \gC_G$,
\begin{align}
    M_{h}\Pi_G(c)P_{h^{-1}} 
    &= M_{h} \left(\frac{1}{|G|}\sum_{g\in G} M_gcP_{g^{-1}}\right) P_{h^{-1}}  \\
    &= \frac{1}{|G|} \sum_{g\in G} M_{h}M_{g} c P_{g^{-1}} P_{h^{-1}} \\
    &= \frac{1}{|G|} \sum_{g\in G} M_{hg} c P_{g^{-1}h^{-1}} \\
    &= \frac{1}{|G|} \sum_{g\in G} M_{hg} c P_{(hg)^{-1}} \\
    &= \frac{1}{|G|} \sum_{h^{-1}r\in G} M_{r} c P_{(r)^{-1}} \\
    &= \frac{1}{|G|} \sum_{r\in hG} M_{r} c P_{(r)^{-1}} \\
    &= \frac{1}{|G|} \sum_{r\in G} M_{r} c P_{(r)^{-1}} \\
    &= \Pi_G(c).
\end{align}
Thus we have $\Pi_G(c)\in \gC_G$. 

As for the other property, $\forall g\in G, c\in\gC_G$, we have $M_gcP_{g^{-1}}=c$, that is $M_gc=cP_{g^{-1}}^{-1}=cP_g$. This follows:
\begin{align}
    \Pi_G(c) 
    &= \frac{1}{|G|}\sum_{g\in G} M_gcP_{g^{-1}} \\
    &= \frac{1}{|G|}\sum_{g\in G} cP_gP_{g^{-1}} \\
    &=  \frac{1}{|G|}\sum_{g\in G} c \\
    &= c.
\end{align}
thus $\forall c\in\gC_G, \Pi_G(c)=c$.

Combining these two properties, \cref{thm:symmetrizer} is proved.

\textbf{An Example for Symmetry Map $\Pi_G$.} We provide an example of $K_0$ of symmetry map $\Pi_G$ here. In \cref{fig:framework}(c), note that $K_0=\{\pi_0,e\}$ and we thus have
\begin{align}
    \Pi_{K_0}(c) &= \frac12\left(M_{\pi_0}cP_{\pi_0^{-1}} + M_ecP_{e^{-1}}\right) \\ 
    &= \frac12 \Bigg(\begin{pmatrix}
        x_1 & x_4 & x_3 & x_2 \\
        -y_1 & -y_4 & -y_3 & -y_2
    \end{pmatrix} + 
     \begin{pmatrix}
        x_1 & x_2 & x_3 & x_4 \\
        y_1 & y_2 & y_3 & y_4
    \end{pmatrix} \Bigg) \\
    &= \begin{pmatrix}
        x_1 & (x_2+x_4)/2 & x_3 & (x_2+x_4)/2 \\
        0 & (y_2-y_4)/2 & 0 & (y_4-y_2)/2
    \end{pmatrix}.
\end{align}

\subsection{Proof of \cref{thm:symmetry}}\label{appx:symmetry_all}
Define $\Xi\triangleq (\Delta_G^{skel},\Delta_G^{attr},\Pi_G)$, where $\Delta_G^{skel}$ is defined in \cref{equ:sym_skel}, $\Delta_G^{attr}$  is defined in \cref{equ:sym_attr}, $\Pi_G$ is defined in \cref{equ:symmetrizer}. 
Here we prove the equivalence between transformed design space by transformations $\Xi$ and $G$-symmetric space. 

In \cref{sec:design_a_robot}, we have shown that any transformed robots by $\Xi$ are $G$-symmetric. Thus we only need to show that any $G$-symmetric robots can be generated by $\Xi$. To prove this, a sufficient condition is the fixing property of $\Xi$, \ie{}, $\Xi$ will leave any $G$-symmetric robots unchanged. Since we have proved the fixing property of $\Pi_G$, here we only need to show that $\Delta_G^{skel}$ and $\Delta_G^{attr}$ also have this property. 

In a $G$-symmetric robot design $D=\{V,E,Z\}$, for any joint $v\in V$ and any joint $u\in \gO_G(v)$,  $v$ and $u$ must have the same number of child joints and same attributes (according to the definition of $G$-symmetric). Assuming $D$ is generated layer by layer and the current design is $D'=(V',E',Z')$, in layer $j$, $\forall v\in V_j$, we have that all joints in $\gO_G(v)$ can choose the same skeleton action and scalar attribute action as $D$. That is $a_v^{skel}=$ AddJoint if $\sum_{(v,u)\in E, (v,u)\notin E'}1 > 0$ else NoChange, and $a_v^{atrr}=z_v\in Z$. Thus the design is left unchanged. 

\subsection{Neighbors of a Subgroup}\label{appx:subgroup_neighbor}
$\forall G'\in \textit{Neighbor}(G)$, we have 
\begin{itemize}
    \item $G'<G$ or $G<G'$;
    \item if $G<G'$ and $\exists H, G<H<G'$, we have $G=H$ or $G'=H$;
    \item if $G'<G$ and $\exists H, G'<H<G$, we have $G=H$ or $G'=H$;
\end{itemize}

\subsection{Derivation of $\Pi_{G,G',\beta}$}\label{appx:derive_beta}
Here we prove that
\begin{align}
    \Pi_{G'}(c) &= \beta_0 \Pi_{G}(c) + (1-\beta_0)\Pi_{G'\shortn G}(c) 
\end{align}
where $\beta_0=|G|/|G'|$ and $G'\shortn G\triangleq \{g|g\in G', g\notin G\}$. 
\begin{align}
    \Pi_{G'}(c)&=\frac{1}{|G|}\sum_{g\in G'} M_gcP_{g^{-1}}  \\
    &= \frac{|G|}{|G'|}\cdot \frac{1}{|G'|}\sum_{g\in G} M_gcP_{g^{-1}} + \frac{|G'|-|G|}{|G'|}\cdot \frac{1}{|G'|-|G|}\sum_{g\in G'-G} M_gcP_{g^{-1}} \\
    &= \frac{|G|}{|G'|} \Pi_{G}(c) + \frac{|G'|-|G|}{|G'|} \Pi_{G'\shortn G}(c) \\
    &= \beta_0 \Pi_{G}(c) + (1-\beta_0)\Pi_{G'\shortn G}(c).
\end{align}


\subsection{A Theorem for $\Pi_{G,G',\beta}$}\label{appx:symmetrizer_beta}

For symmetry map $\Pi_{G,G',\beta}$ defined in \cref{equ:symmetrizer_beta}, we also have a similar theorem as \cref{thm:symmetrizer} in the following:
\begin{theorem}\label{thm:symmetrizer_beta}
The projected vector values $\Pi_{G,G',\beta}$ defined in \cref{equ:symmetrizer_beta} are $G$-symmetric. And if $c$ is already $G'$-symmetric, $\Pi_{G,G',\beta}(c)=c$.
\end{theorem}

For $\forall h\in G, c\in \gC_G$,
\begin{align}
    M_{h}\Pi_{G,G',\beta}P_{h^{-1}} 
    &= M_{h} \left(\beta \Pi_{G}(c) + (1-\beta)\Pi_{G'\shortn G}(c)\right) P_{h^{-1}}  \\
    &= M_{h} \left(\frac{\beta}{|G'|}\sum_{g\in G} M_gcP_{g^{-1}} + \frac{(1-\beta)}{|G'|-|G|}\sum_{g\in G'-G} M_gcP_{g^{-1}} \right) P_{h^{-1}}  \\
    &= \frac{\beta}{|G'|}\sum_{g\in G} M_{h}M_gcP_{g^{-1}}P_{h^{-1}} + \frac{(1-\beta)}{|G'|-|G|}\sum_{g\in G'-G} M_{h}M_gcP_{g^{-1}}P_{h^{-1}} \\
    &= \frac{\beta}{|G'|}\sum_{g\in G} M_{hg}cP_{(hg)^{-1}} + \frac{(1-\beta)}{|G'|-|G|}\sum_{g\in G'-G} M_{hg}cP_{(hg)^{-1}} \\
    &= \frac{\beta}{|G'|}\sum_{h^{-1}r\in G} M_{r}cP_{r^{-1}} + \frac{(1-\beta)}{|G'|-|G|}\sum_{h^{-1}r\in G'-G} M_{r}cP_{r^{-1}} \\
    &= \frac{\beta}{|G'|}\sum_{r\in hG} M_{r}cP_{r^{-1}} + \frac{(1-\beta)}{|G'|-|G|}\sum_{r\in h(G'-G)} M_{r}cP_{r^{-1}} \\
    &= \frac{\beta}{|G'|}\sum_{r\in G} M_{r}cP_{r^{-1}} + \frac{(1-\beta)}{|G'|-|G|}\sum_{r\in G'-G} M_{r}cP_{r^{-1}} \label{equ:prove_beta}\\
    &= \beta \Pi_{G}(c) + (1-\beta)\Pi_{G'\shortn G}(c) \\
    &= \Pi_{G,G',\beta}(c).
\end{align}
Here in \cref{equ:prove_beta}, $hG=G$ is a basic property of group theory, and we prove $h(G'\shortn G)=G'\shortn G$. Any element in $h(G'\shortn G)$ can be represented by $hg$ where $g\in G'\shortn G$. If $hg\notin G'\shortn G$, we have $hg\in G$. Assuming $g_1\in G$ such that $hg=g_1$, we have $g=h^{-1}g_1\in G$, which leads to a contradiction and thus $hg\in G'\shortn G$. On the other hand, $\forall g_2 \in G'\shortn G$, we have $g_2=h(h^{-1}g_2)$ and only need to show that $h^{-1}g_2\in G'\shortn G$. Otherwise, assuming $h^{-1}g_2\notin G'\shortn G$, we have $h^{-1}g_2\in G$ and thus $g_2\in G$, which also leads to a contradiction and thus $g_2\in h(G'\shortn G)$. In conclusion, $h(G'\shortn G)=G'\shortn G$

Therefore, we have $\Pi_{G,G',\beta}(c)\in \gC_{G}$. 

As for the fixing property, $\forall g\in G, c\in\gC_{G'}$, we have $M_gcP_{g^{-1}}=c$, that is $M_gc=cP_{g^{-1}}^{-1}=cP_g$. This follows:
\begin{align}
    \Pi_{G,G',\beta}(c) 
    &= \frac{\beta}{|G'|}\sum_{g\in G} M_gcP_{g^{-1}} + \frac{(1-\beta)}{|G'|-|G|}\sum_{g\in G'-G} M_gcP_{g^{-1}} \\
    &= \frac{\beta}{|G'|}\sum_{g\in G} cP_gP_{g^{-1}} + \frac{(1-\beta)}{|G'|-|G|}\sum_{g\in G'-G} cP_gP_{g^{-1}} \\
    &= \frac{\beta}{|G'|}\sum_{g\in G} c + \frac{(1-\beta)}{|G'|-|G|}\sum_{g\in G'-G} c \\
    &= c.
\end{align}
thus $\forall c\in\gC_{G'}, \Pi_{G,G',\beta}(c)=c$.



\section{Experiment Details}\label{appx:exp}

\subsection{Details of the Tasks}\label{appx:task_detail}
\begin{figure*}[htp]
\centering
\includegraphics[width=\linewidth]{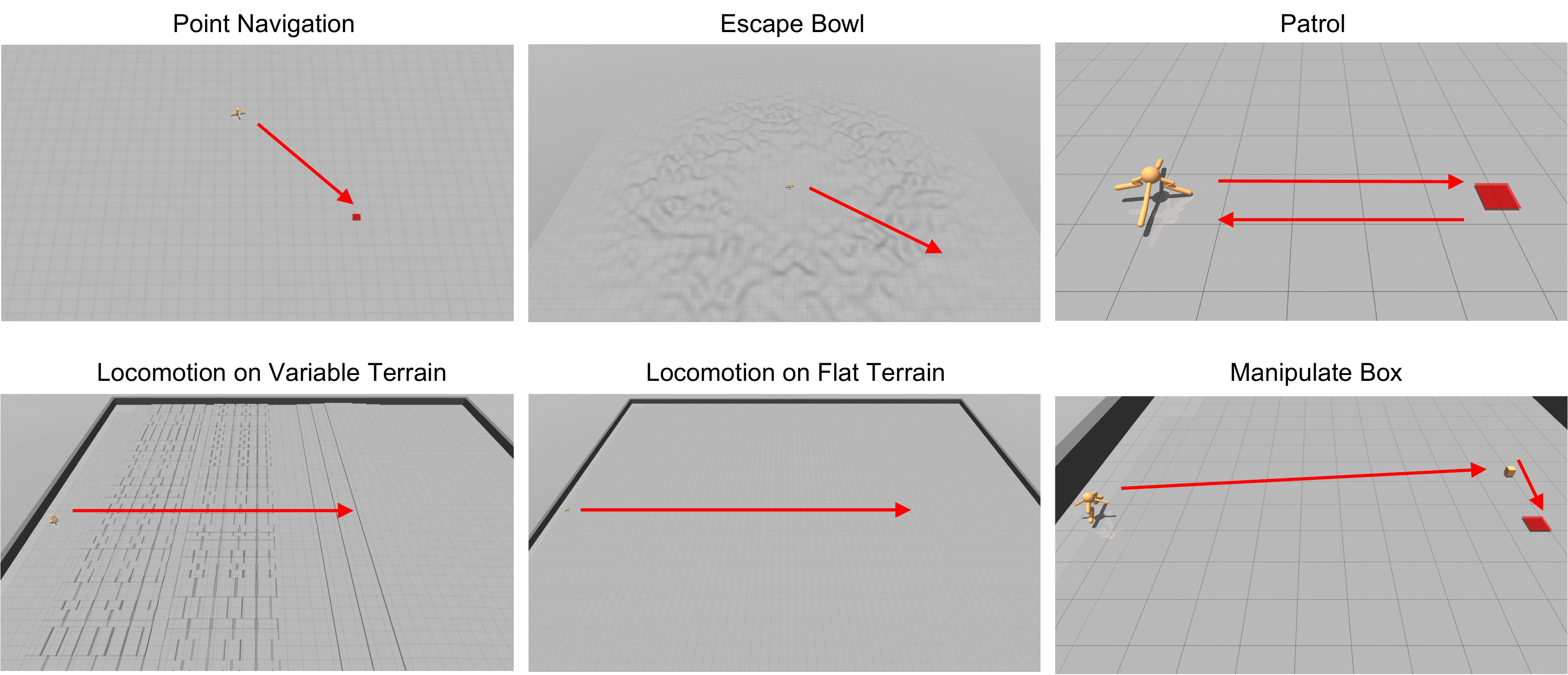}
\caption{Visualization of the training tasks adapted from \citet{gupta2021embodied}}\label{fig:tasks}
\end{figure*}

We run experiments on six MuJoCo~\cite{todorov2012mujoco} tasks adapted from ~\citet{gupta2021embodied}. These tasks can be categorized into 3 domains to test agility (\patrol{}, \pointnav{}), stability (\escape{},\locomotionvt{},\locomotionft{}), and manipulation (\manipulationbox{}) abilities of the designed robots. The detailed descriptions of these tasks are listed below. 

\textbf{\pointnav{}.} The agent is generated at the center of a $100\times100\ m^{2} $ flat arena and needs to reach a random goal (red square in \cref{fig:tasks}) in this arena. The ability to move in any specified direction quickly leads to success in this task. At each time step, the agent receives a reward $r_{t}$ shown below,
\begin{align*}
    r_{t} =w_{ag} d_{ag}-w_{c}\left \| a \right \| ^{2} 
\end{align*}
where $d_{ag}$ is the geodesic distance difference in the current time step and previous time step between the agent and the goal, $w_{ag}=100$, $w_{c}=0.001$, $w_c$ is a penalty term for action $a$. 

\textbf{\escape{}.} Generated at the center of a bowl-shaped terrain surrounded by small hills, the agent has to escape from the hilly region. This task requires the agent to maximize the geodesic distance from the initial location while going through a random hilly terrain. At each time step, the agent receives a reward $r_{t}$ shown below,
\begin{align*}
    r_{t} =w_{d} d_{as}-w_{c}\left \| a \right \| ^{2} 
\end{align*}
where $d_{as}$ is the geodesic distance difference in the current time step and previous time step between the agent and the initial location, $w_{d}=1$, $w_{c}=0.001$.

\textbf{\patrol{}.} In the task, the agent is required to run back and forth between two target locations at a distance of 10 meters along the x-axis. Quick change of direction when the goal (red square in \cref{fig:tasks}) alters and rapid movement leads to the success of this task. The reward function is similar to the point navigation task. Additionally, we flip the goal location and provide the agent a sparse reward of 10 as it is within 0.5$m$ from the goal location.

\textbf{\locomotionvt{}.} At the beginning of an episode, an agent is generated on the one end of a $100\times100\ m^{2} $ square arena. By randomly sampling a sequence of obstacles from a uniform distribution over a predefined range of parameter values, we can build a brand new terrain in each episode. While the length of flat segments in variable terrain $l\in [1,3]m$ along the desired direction of motion, the length of obstacle segments $l\in [4,8]m$. We primarily utilize 3 types of obstacles. 1) Hills, which is parameterized by the amplitude $a$ of sin wave in which $a\in [0.6,1.2]m$. 2) Steps, a sequence of 8 steps of height 0.2m. 3) Rubble, a sequence of random bumps (small hills) created by clipping a repeating triangular sawtooth wave at the top and height $h$ of each bump clip samples from $[0.2,0.3]m$ stochastically. The goal of the agent is to maximize forward displacement over an episode and this environment is quite challenging for the agent to perform well.

\textbf{\locomotionft{}.} Similar to locomotion on variable terrain task, an agent is initialized on the one end of a $150\times150\ m^{2} $ square arena and aims at maximizing forward displacement over an episode.

\textbf{\manipulationbox{}.} In a $60\times40\ m^{2} $ arena similar to variable terrain, the agent is required to move a box (small cube with $0.2m$ shown in \cref{fig:tasks}) from the initial position to the target place (red square). Both the initial box location and final target location are randomly chosen with constraints that lead to a further path to the destination in each episode.

\subsection{Implementation of \name{}}
We implement \name{} based on \tact{}~\cite{yuan2021transform2act}, which uses GNN-based~\cite{scarselli2008graph, bruna2013spectral, kipf2016semi} control policies. GNN-based policies can deal with variable input sizes across different robot designs by sharing parameters between joints. This property allows us to share policies across all designed robots. Note that our method is general and can be combined with any other network structures used in modular RL, \eg{}, message passing networks~\cite{huang2020one} and Transformers~\cite{dong2022solar, kurin2020morphenus}. 

However, this sharing also brings negative impacts, \eg{}, joints in similar states will choose similar actions, which may severely hinder performance. To solve this problem, \tact{} proposed to add a joint-specialized MLP (JSMLP) after the GNNs. We follow this setting for fair comparisons. 

For design policy and control policy learning, we use Proximal Policy Optimization (PPO)~\cite{schulman2017PPO}, a standard policy gradient method~\cite{williams1992PG} for optimizing these two policies.

Here we provide the hyperparameters needed to replicate our experiments in \cref{tab:hyperpara}, and we also include our codes in the supplementary.

\begin{table}[]
\centering
\caption{Hyperparameters of \name{} and \tact{}.}\label{tab:hyperpara}
\begin{tabular}{ll}
\hline
Hyperparameters                                                   & Value      \\ \hline
\multicolumn{1}{l|}{Skeleton Design Stage Time Steps $N^{skel}$}  & 5          \\
\multicolumn{1}{l|}{Attribute Design Stage Time Steps $N^{attr}$} & 1          \\
\multicolumn{1}{l|}{GNN Layer Type}                               & GraphConv  \\
\multicolumn{1}{l|}{JSMLP Activation Function}                    & Tanh       \\
\multicolumn{1}{l|}{GNN Size}                                     & (64,64,64) \\
\multicolumn{1}{l|}{JSMLP Size}                                   & (128,128)  \\
\multicolumn{1}{l|}{Policy Learning Rate}                         & 5e-5       \\
\multicolumn{1}{l|}{Value Learning Rate}                          & 3e-4       \\
\multicolumn{1}{l|}{PPO Clip}                                     & 0.2        \\
\multicolumn{1}{l|}{PPO Batch Size}                               & 50000      \\
\multicolumn{1}{l|}{PPO MiniBach Size}                            & 2048       \\
\multicolumn{1}{l|}{PPO Iterations Per Batch}                     & 10         \\
\multicolumn{1}{l|}{Training Epochs}                              & 1000       \\
\multicolumn{1}{l|}{Discount Factor $\gamma$}                     & 0.995      \\
\multicolumn{1}{l|}{GAE $\lambda$}                                & 0.95       \\
\multicolumn{1}{l|}{Subgroup Exploration Rate $\epsilon$}         & 0.01       \\ \hline
\end{tabular}
\end{table}

Experiments are carried out on NVIDIA GTX 2080 Ti GPUs. Taking \pointnav{} as an example, \name{} requires approximately 10G of RAM and 4G of video memory and takes about 36 hours to finish 50M timesteps of training.

\subsection{Details of Baselines}\label{appx:tact_detail}
\textbf{\tact{}.} We use the official implementation of \tact{}, where all networks and optimizations are implemented with PyTorch~\cite{paszke2019pytorch}. The GNN layers are GraphConv~\cite{morris2019graphconv} implemented in PyTorch Geometric package~\cite{fey2019geometric}. All policies are optimized with PPO~\cite{schulman2017PPO} with generalized advantage estimation (GAE)~\cite{schulman2015gae}. The authors searched the hyperparameters and we also list the selected values in \cref{tab:hyperpara}. We also removed the initial design of \tact{} to avoid any prior knowledge.

\textbf{\originalant{}.} To show the strength of robot design, we also compare \name{} with \originalant{}, which is the human-designed robot Ant\footnote[2]{\url{https://github.com/openai/gym/blob/master/gym/envs/mujoco/assets/ant.xml}} using expert knowledge from OpenAI. We directly load the XML file and skip the Design Stage. The Control Stage and optimization are the same as ours. We run 50M time steps for it and show the best performance.

\subsection{Details of Ablations}\label{appx:ablation_detail}

\textbf{\nameUnstruct{}.} The $\textit{Neighbor}(G_i)$ function at iteration $i$ is set to contain all subgroups and other components are the same as \name{}.

\textbf{\nameKfive{}.} Divide the interval between two adjacent subgroups into $K=5$ parts and keep others the same as \name{}.

\textbf{\nameKone{}.} Do not divide the interval between two adjacent subgroups and only use original group structures to define $\textit{Neighbor}(G_i)$ function at iteration $i$.

\section{Extra Results}\label{appx:extra_results}
\subsection{Combine \name{} with Other Robot Design Method}

\begin{table}[]
\centering
\caption{Training performance of \name{} based on different base algorithms. }\label{tab:other_impl}
\begin{tabular}{@{}lllll@{}}
\toprule
                        & \nge{}    & \tact{} & \name{}+\nge{}                & \name{}+\tact{}      \\ \midrule
\pointnav{}               & $1131.50$ $\pm$458.45 & 1618.10$\pm$1022.01       & \textbf{4729.00$\pm$835.62} & 4262.78$\pm$738.17          \\
\escape{}                         & 8.65$\pm$1.88      & 32.37$\pm$13.62           & 15.55$\pm$3.69              & \textbf{88.61$\pm$13.23}    \\
\patrol{}                         & 1120.30$\pm$425.89 & 1995.95$\pm$709.07        & 3104.67$\pm$1082.03         & \textbf{3116.47$\pm$801.43} \\
\locomotionvt{}     & 170.85$\pm$48.05   & 443.22$\pm$74.72          & 408.65$\pm$74.95            & \textbf{1204.01$\pm$96.16}  \\
\locomotionft{}     & 238.65$\pm$86.75   & 1067.16$\pm$463.55        & 835.75$\pm$366.25           & \textbf{2438.26$\pm$297.09} \\
\manipulationbox{}                 & 1061.90$\pm$541.47 & 1073.11$\pm$467.38        & \textbf{1793.00$\pm$27.80}  & 1604.27$\pm$137.72          \\ \bottomrule
\end{tabular}
\end{table}

Our method is a plug-and-play module that can be utilized in other robot design methods. Here we provide experimental results of combining our method (\name{}) with \nge{}~\citep{wang2019neural}, which is an ES-based robot design method. We report the results in \cref{tab:other_impl}. Here \name{}+\nge{} is the implementation of \name{} based on \nge{}, and \name{}+\tact{} is our original implementation of \name{} based on \tact{}, denoted by `\nameKthree{}' in our paper. All runs are conducted with 3 random seeds and each data item in the table is formatted as `mean$\pm$std'. As shown in \cref{tab:other_impl}, \name{}+\nge{} outperforms vanilla \nge{} in all tasks and is even better than our original implementation \name{}+\tact{} in two tasks. The performance improvement of \name{}+BaseAlgo over BaseAlgo (BaseAlgo $\in$ \{\nge{}, \tact{}\}) showcases the generality of \name{}.

\subsection{Results of \name{} on the Tasks Used in \tact{{}}}
\begin{table}[]
\centering
\caption{Training performance of \name{} compared with \tact{} on its original tasks. }\label{tab:ori_task}
\begin{tabular}{@{}llll@{}}
\toprule
              & Swimmer              & 2D Locomotion            & Gap Crosser              \\ \midrule
\tact{} & 607.50$\pm$89.02         & \textbf{3329.00$\pm$2094.56} & 1352.20$\pm$558.07           \\
\name{}          & \textbf{975.50$\pm$9.10} & 3194.00$\pm$1695.06          & \textbf{1824.43$\pm$1322.30} \\ \bottomrule
\end{tabular}
\end{table}

For a complete comparison, we also provide extra results on the tasks used in \tact{} \citep{yuan2021transform2act}. We show the final performance comparison between our method \name{} and \tact{} in \cref{tab:ori_task}. Here \name{} is our original implementation of \name{} based on \tact{}, denoted by `\nameKthree{}' in our paper. All runs are conducted with 3 random seeds and each data item in the table is formatted as `mean±std'. The results that \name{} outperforms \tact{} in most tasks further validate the strength of our method. Also, please note that the `3D Locomotion' task in their paper is similar to `\locomotionft{}' in \cref{tab:other_impl}, thus we omit this task here. The reported results of Transform2Act are based on their released code.

\begin{figure*}[ht]
\centering
\includegraphics[width=0.9\linewidth]{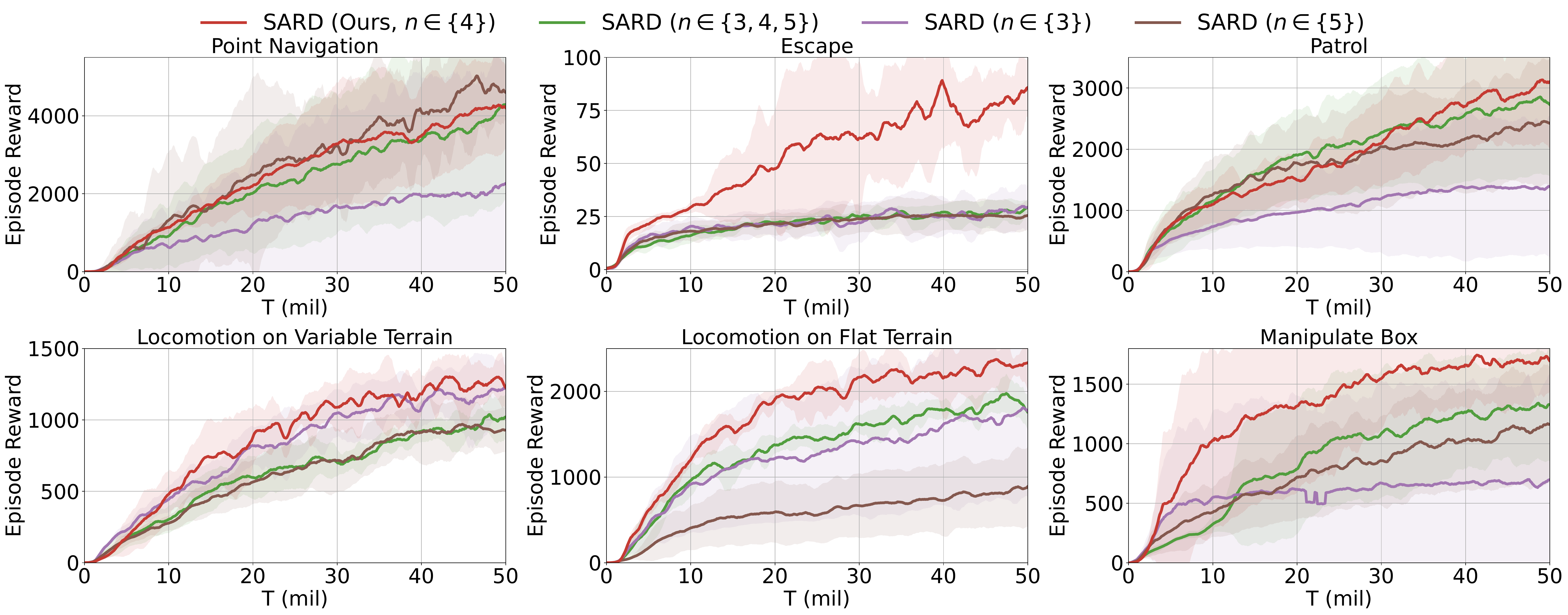}
\vspace{-1em}
\caption{Hyperparameter search of $\text{Dih}_n$.}\label{fig:group_size}
\vspace{-1em}
\end{figure*}

\subsection{Different Dihedral Groups}
In this paper, we use the subgroups of the dihedral group $\text{Dih}_n$ to represent various symmetries and in \cref{sec:exp}, we set the hyperparameter $n$ to 4. Here we conduct a hyperparameter search to verify this choice. The result is shown in \cref{fig:group_size}. 

\name{}($n\in\{3\}$), \name{}($n\in\{4\}$), \name{}($n\in\{5\}$) are \name{} with different dihedral group, \ie{}, $\text{Dih}_3$, $\text{Dih}_4$, $\text{Dih}_5$, respectively. As for \name{}($n\in\{3,4,5\}$), we use these three groups simultaneously by regarding group elements with the same matrix representations as neighbors. 

\name{} outperforms all others in most tasks and is only a little worse than \name{}($n\in\{5\}$ in \pointnav{} task. This result validates our hyperparameter choice.


\section{Discussions of Dihedral Groups}
In this paper, we use Dihedral groups to describe the symmetry of robots mainly for two reasons. (1) The Dihedral groups are generally enough to represent a wide range of symmetries of the robot's morphologies. This is because the Dihedral groups are generated by basic reflectional and rotational symmetries, which can describe the characteristics of most effective robot morphologies. Besides, related works from biology \citep{savriama2011symmetry, pappas2021uncanny, graham2010fluctuating} also use the Dihedral group as an effective tool to study the symmetry of real-world creatures. (2) Using larger groups may bring in extra learning complexity and lead to poor performance, even though larger groups can contain more symmetries. In general, the Dihedral group is a good trade-off between expressiveness and complexity

\section{Limitations}

In this paper, we use the subgroups of dihedral groups $\text{Dih}_n$ to represent a wide range of symmetries while still avoiding extra learning complexity. Our method has shown superior efficiency, but the dihedral group is a 2D symmetry group and only contains transformations in the xy-plane. Perhaps because of the influence of gravity, dihedral groups are enough for representing the symmetries of most real-world creatures. However, it is still worthwhile to explore 3D symmetry groups~\cite{savriama2011symmetry} in the virtual robot design problem, which might be a promising future work.

In addition, although the idea of symmetry can be applied to a wide range of tasks, it may not be suitable for tasks that do not require symmetry, such as single-arm robotic manipulation tasks where we need to design a robot arm as well as its gripper for a particular manipulation task. Intuitively, efficient designs are mostly asymmetric in these situations, and a symmetry constraint might prevent the arm and manipulator from operating in a more effective way, thus hindering training.




\end{document}